\documentclass{article}
\ifdefined\pdfminorversion\pdfminorversion=7\fi
\PassOptionsToPackage{numbers,sort&compress}{natbib}
\usepackage[preprint]{neurips_2026}
\usepackage[utf8]{inputenc}
\usepackage[T1]{fontenc}
\usepackage{amsmath,amssymb,mathtools,amsthm}
\usepackage{graphicx,float}
\usepackage{placeins}
\usepackage{wrapfig}
\usepackage{ragged2e}
\usepackage{needspace}
\usepackage{booktabs,tabularx,colortbl,makecell,multirow}
\usepackage{caption,subcaption}
\usepackage{nicefrac,microtype,xspace,xcolor}
\usepackage{url}
\usepackage{marvosym}
\usepackage{hyperref}
\usepackage[capitalize,noabbrev]{cleveref}
\newcolumntype{Y}{>{\centering\arraybackslash}X}
\definecolor{linecolor}{gray}{.65}
\definecolor{kaiming-green}{RGB}{57,181,74}
\newcommand{\gr}[1]{\textcolor{gray}{#1}}
\newcommand{\modelname}{FlexPart\xspace}

\graphicspath{{figures/}}
\hypersetup{
  colorlinks=true,
  linkcolor=red,
  filecolor=magenta,
  urlcolor=magenta,
  citecolor=kaiming-green,
  pdftitle={A Unified Framework for Image-to-3D Part Generation via Variable Granularity},
  pdfauthor={Qiwen Gu, Xin Zhou, Xiwu Chen, Zhongyang Zhu, Zhuo Zhang, Junqiao Zhao}
}
\date{}

\title{A Unified Framework for Image-to-3D Part Generation via Variable Granularity}

\author{%
  \textbf{Qiwen Gu}\textsuperscript{1,2,\textdaggerdbl}\quad
  \textbf{Xin Zhou}\textsuperscript{2,3,\textdaggerdbl}\quad
  \textbf{Xiwu Chen}\textsuperscript{2,\Letter,\textdagger} \\
  \textbf{Zhongyang Zhu}\textsuperscript{2}\quad
  \textbf{Zhuo Zhang}\textsuperscript{2}\quad
  \textbf{Junqiao Zhao}\textsuperscript{1,\Letter} \\[4pt]
  \normalfont\textsuperscript{1}Tongji University\quad
  \textsuperscript{2}Chongqing Afari Intelligent Drive Co., Ltd \\
  \normalfont\textsuperscript{3}Huazhong University of Science and Technology \\[4pt]
  {\small\hypersetup{urlcolor=black}%
    \href{mailto:2432178@tongji.edu.cn}{\texttt{2432178@tongji.edu.cn}},\quad
    \href{mailto:xzhou03@hust.edu.cn}{\texttt{xzhou03@hust.edu.cn}}}
}

\begin{document}

\maketitle
\begingroup
\renewcommand{\thefootnote}{\fnsymbol{footnote}}
\footnotetext[2]{Project leader. $^{\text{\Letter}}$~Corresponding author.}
\footnotetext[3]{Work done during an internship at Chongqing Afari Intelligent Drive Co., Ltd.}
\vspace{-10pt}
\endgroup

\begin{figure}[H]
    \centering
    \includegraphics[width=0.90\linewidth]{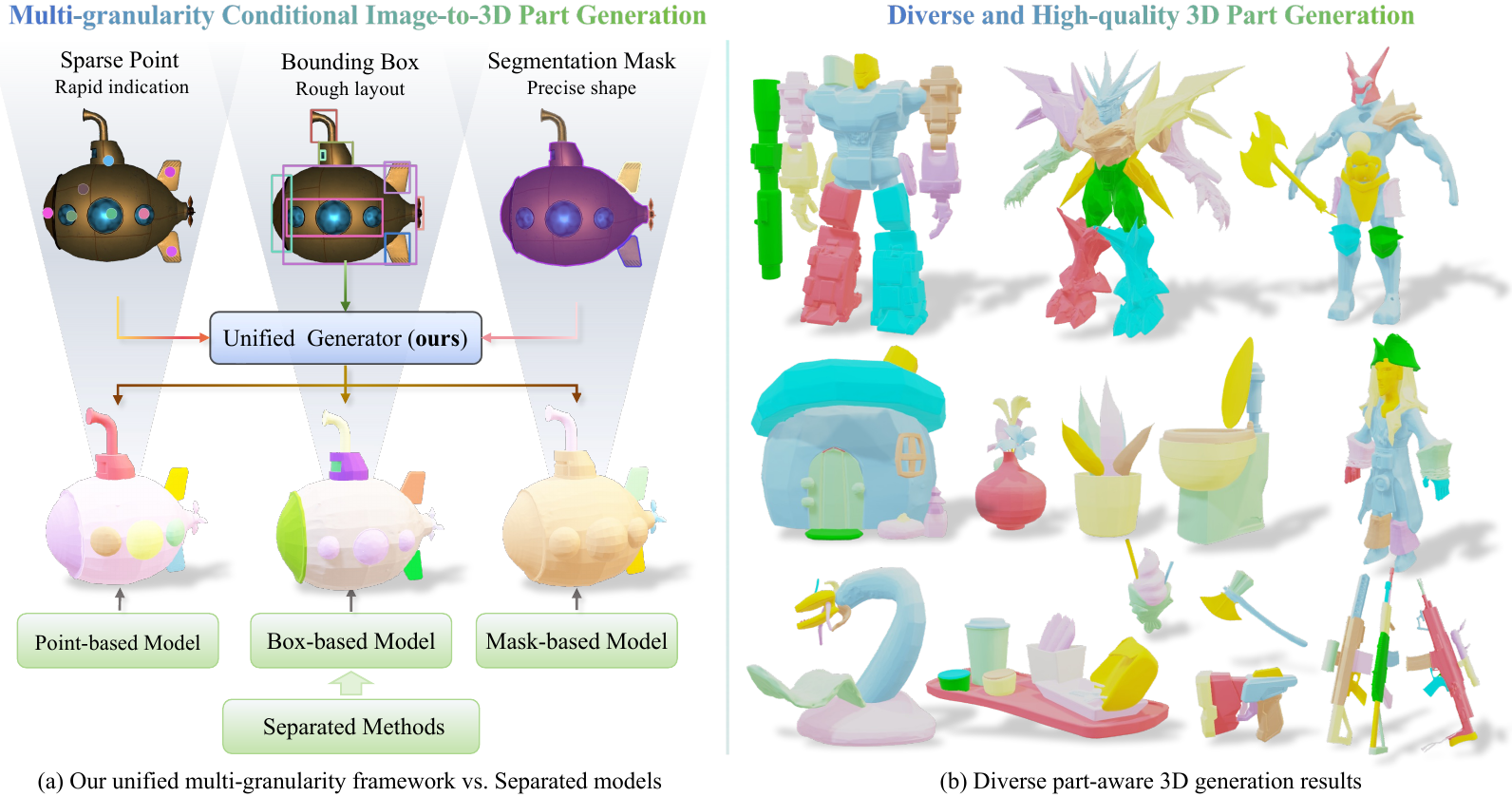}
    \caption{
        (a) Our unified framework integrates variable-granularity geometric prompts from sparse points to dense masks into one model for flexible part control. 
        (b) The proposed method produces high-fidelity, semantically coherent 3D parts across diverse object categories.
    }
    \label{fig:intro}
\end{figure}

\begin{abstract}
  Controllable 3D part generation is pivotal for digital asset creation, yet existing methods lack the flexibility to accommodate diverse user workflows requiring varying levels of control. In this paper, we introduce \modelname, a unified image-to-3D part generation framework that supports variable-granularity geometric conditions within a single model. By seamlessly integrating heterogeneous inputs, ranging from sparse points to dense bounding boxes and masks, via gated adaptive modulation, \modelname unlocks a powerful cross-modal synergy. We also propose an asymmetric geometric guidance mechanism that leverages strong spatial priors to align representations across granularities, improving geometric consistency and facilitating inter-task synergy. Under mask conditioning on PartVerseXL, \modelname reduces Part CD by 16.6\% relative to TripoSG + SAM3D and by 10.5\% relative to OmniPart, while maintaining competitive whole-mesh quality.
  The code can be found at \url{https://github.com/Gucci233/FlexPart}
\end{abstract}

\section{Introduction}

Perceiving objects as compositions of semantically coherent parts is essential for various tasks such as animation, simulation, and interactive editing. Unlike holistic 3D generation methods that generate whole entities~\cite{zhang2024clay,wu2024direct3d,li2025triposg,xiang2025structured}, image-to-3D part generation decomposes visual inputs into distinct, spatially assemblable, structurally detachable, and image-consistent 3D components.

However, this task presents significant challenges due to the inherent ambiguity of part decomposition. For example, the same object can be segmented at various granularities, and users may require different structural specifications depending on their goals. A straightforward solution is to project 2D segmentation masks~\cite{yang2023sam3d} onto generated 3D meshes. However, such decoupled pipelines can introduce geometric inconsistencies during the lifting process. To address this limitation, recent approaches have integrated part control directly. Some methods rely on dense geometric prompts, such as pixel-wise 2D segmentation masks~\cite{chen2025partgen,yang2025omnipart,dong2025one} or explicit bounding box layouts~\cite{ding2025fullpart}. These inputs provide strong spatial constraints, enabling well-structured part generation. In contrast, other methods explore lighter forms of supervision, such as high-level prompts like part counts~\cite{lin2025partcrafter} or implicit volumetric partitioning~\cite{tang2025efficient}. These approaches reduce annotation effort and improve usability but often sacrifice precise spatial control.

While effective within their respective domains, existing methods are typically tailored to a single prompt granularity, as illustrated in Fig.~\ref{fig:intro}(a). This specialization not only limits user flexibility by hindering seamless transitions between fine-grained control via dense masks and lightweight interaction via sparse points, but also overlooks the inherent structural relationships across different prompt modalities. Such rigidity is suboptimal, as prompts of varying granularities provide complementary information that could be jointly leveraged. This raises a fundamental question: \textit{can we develop a unified framework that accommodates variable-granularity inputs for flexible part generation, while exploiting their cross-modal synergy to enhance quality?}

Driven by this motivation, we propose \textbf{\modelname}, a unified framework for controllable 3D part generation. As shown in Fig.~\ref{fig:intro}(a), \modelname integrates sparse points, bounding boxes, and masks into a single Diffusion Transformer (DiT) via gated adaptive modulation. Joint training improves generation quality across prompt granularities. To further amplify this synergy, we introduce an Asymmetric Geometric Guidance mechanism. During training, it employs differentiable optimal transport to let strong spatial priors (masks/boxes) explicitly supervise the ambiguous representations of sparse points. The resulting model supports different prompt types at inference without retraining.

By consolidating diverse geometric constraints, \modelname establishes a unified paradigm for versatile part creation with variable granularity. Under mask conditioning on PartVerseXL, \modelname reduces Part CD by 16.6\% relative to TripoSG + SAM3D and by 10.5\% relative to OmniPart. Thanks to the learned cross-modal synergy, it remains highly competitive even with significantly sparser prompts. Ultimately, \modelname delivers superior part-level generation quality (Fig.~\ref{fig:intro}(b)) while preserving strong whole-mesh reconstruction fidelity.

Our contributions are summarized as follows: 
\textbf{1)} We propose \modelname, the first unified framework integrating point, box, and mask prompts into a single DiT to unlock cross-modal synergy for controllable 3D part generation;
\textbf{2)} We devise an Asymmetric Geometric Guidance mechanism that effectively transfers structural knowledge from dense priors to enhance the spatial reasoning of sparse inputs;
\textbf{3)} We demonstrate that our method achieves superior generation quality and geometric consistency compared to existing state-of-the-art approaches across diverse input modalities.

\section{Related Work}

\textbf{General 3D Mesh Generation.}
Early 3D generation methods relied on optimization-based pipelines, using Score Distillation Sampling (SDS) to distill 2D priors into implicit 3D representations~\cite{poole2022dreamfusion,lin2023magic3d,wang2023prolificdreamer,voleti2024sv3d}. While pioneering, these approaches often suffer from slow convergence and geometric artifacts. Feed-forward Large Reconstruction Models (LRMs)~\cite{hong2023lrm,tang2024lgm} improve efficiency by directly predicting 3D representations from images.
Native 3D generative models use compact latent representations~\cite{zhang20233dshape2vecset,zhang2024clay,chen2025dora} to support efficient diffusion training. Recent methods~\cite{xiang2025structured,yang2024hunyuan3d,zhao2025hunyuan3d,li2025triposg} further improve geometric fidelity through structured sparse latents or flow matching. These object-level generators typically produce holistic meshes without explicit part decomposition, limiting part-level interaction and editing.
Complementary work connects 3D analysis and generation. ROAD transfers discriminative semantic and structural priors to a generative model through reciprocal-objective alignment~\cite{luo2026road}, while HERMES and HERMES++ jointly model scene understanding and future geometry generation in autonomous driving~\cite{zhou2025hermes,zhou2026hermespp}. Our work focuses on joint learning across geometric prompt granularities for controllable object-part generation.

\textbf{3D Shape Analysis and Part Segmentation.}
Point-cloud analysis has advanced through efficient state-space backbones such as PointMamba~\cite{liang2024pointmamba} and model adaptation methods, including dynamic adapters with prompt tuning~\cite{zhou2024dapt}, spectral-domain fine-tuning~\cite{liang2025pointgst}, and input-aware parameter adaptation for 3D scenes~\cite{liu2026pointtpa}. Here, learned prompts used for model adaptation differ from the user-provided geometric prompts considered in our work. Language-driven approaches such as ShapeLLM~\cite{qi2024shapellm} and Kestrel~\cite{ahmed2025kestrel} further support object understanding and part-aware grounding.
For part decomposition, supervised methods~\cite{mo2019partnet,qi2017pointnet} analyze existing shapes, while other approaches lift 2D foundation model priors, such as SAM~\cite{kirillov2023segment} and GLIP~\cite{li2022grounded}, to 3D~\cite{yang2023sam3d,abdelreheem2023satr,kim2024partstad,liu2024part123,thai20243,zhong2024meshsegmenter}. Recent methods learn open-world 3D feature fields or native segmentation networks~\cite{yang2024sampart3d,liu2025partfield,ma2025p3}. Conventional segmentation primarily identifies regions of existing surfaces. HoloPart~\cite{yang2025holopart} extends this setting by completing initially segmented regions into full 3D parts using a generative model. These pipelines rely on input geometry, whereas our task is to generate complete parts directly from an image and geometric prompts.

\textbf{Part-level 3D Generation.}
Recent generative models synthesize objects as compositions of semantic components, with different ways of specifying part structure. Mask- and layout-conditioned methods~\cite{dong2025one,yang2025omnipart,ding2025fullpart} provide explicit spatial control, although precise masks or layouts can require substantial annotation effort. AutoPartGen~\cite{chen2025autopartgen} generates parts autoregressively and supports conditioning on images, part masks, or existing 3D objects. In a related programmatic reconstruction setting, MeshCoder~\cite{dai2025meshcoder} uses an LLM to translate point clouds into editable, part-structured mesh code.
Methods such as PartCrafter~\cite{lin2025partcrafter} and PartPacker~\cite{tang2025efficient} internalize part decomposition within the generative process without requiring explicit geometric prompts. This simplifies interaction but offers less direct control over spatial part boundaries. A related challenge arises in count-controlled video generation, where NUMINA~\cite{sun2026numina} uses countable latent layouts to improve numerical alignment, illustrating the value of spatial structure for abstract count-based control. Our framework unifies point, box, and mask prompts within one model, allowing users to choose the level of spatial control while benefiting from joint multi-granularity training.

\section{Preliminaries}

\textbf{Optimal Transport.}
Optimal Transport (OT) provides a principled geometric framework for modeling soft correspondences between discrete sets. 
Given two sets of features with cardinalities $N$ and $M$, OT aims to estimate a transport plan 
$\mathbf{P} \in \mathbb{R}_{+}^{N \times M}$, 
where each entry $P_{ij}$ represents the mass transported from element $i$ to element $j$.

Standard OT constrains $\mathbf{P}$ to lie within the transport polytope 
$\mathcal{U}(\mathbf{r}, \mathbf{q})$,
defined by marginal constraints:
\begin{equation}
\mathcal{U}(\mathbf{r}, \mathbf{q})
=
\left\{
\mathbf{P} \in \mathbb{R}_{+}^{N \times M}
\;\middle|\;
\mathbf{P}\mathbf{1}_M = \mathbf{r}, \;
\mathbf{P}^\top \mathbf{1}_N = \mathbf{q}
\right\},
\end{equation}
where $\mathbf{1}$ denotes a vector of ones, and $\mathbf{r}$ and $\mathbf{q}$ are the source and target mass vectors, respectively.

In high-dimensional matching problems, the entropically regularized formulation is widely adopted to enable efficient and differentiable optimization:
\begin{equation}
\mathbf{P}^*
=
\arg\min_{\mathbf{P} \in \mathcal{U}(\mathbf{r}, \mathbf{q})}
\langle \mathbf{P}, \mathbf{C} \rangle
-
\varepsilon H(\mathbf{P}),
\end{equation}
where $\mathbf{C}$ and $\varepsilon$ denote the cost matrix and regularization parameter, respectively. The entropy regularizer is defined as:
\begin{equation}
H(\mathbf{P})
=
-\sum_{i,j} P_{ij} \log P_{ij}.
\end{equation}
The optimal solution admits the factorized form
$\mathbf{P} = \mathrm{diag}(\mathbf{u}) \mathbf{K} \mathrm{diag}(\mathbf{v})$,
with $\mathbf{K} = \exp(-\mathbf{C}/\varepsilon)$,
and can be efficiently computed using the Sinkhorn--Knopp algorithm.

\textbf{Partial Optimal Transport.} Standard OT enforces strict mass conservation, requiring all elements to be matched, which is ill-suited for scenarios with occlusions or noise. Partial OT addresses this by introducing a dustbin mechanism to absorb unmatched mass.
Conceptually, the transport matrix is augmented to
$\bar{\mathbf{P}} \in \mathbb{R}^{(N+1) \times (M+1)}$
allowing elements with no confident correspondence to be assigned to the dustbin row or column. This relaxation handles partial and ambiguous assignments.

\graphicspath{{figures/}} 
\begin{figure*}[ht]
  \centering
  \includegraphics[width=\linewidth]{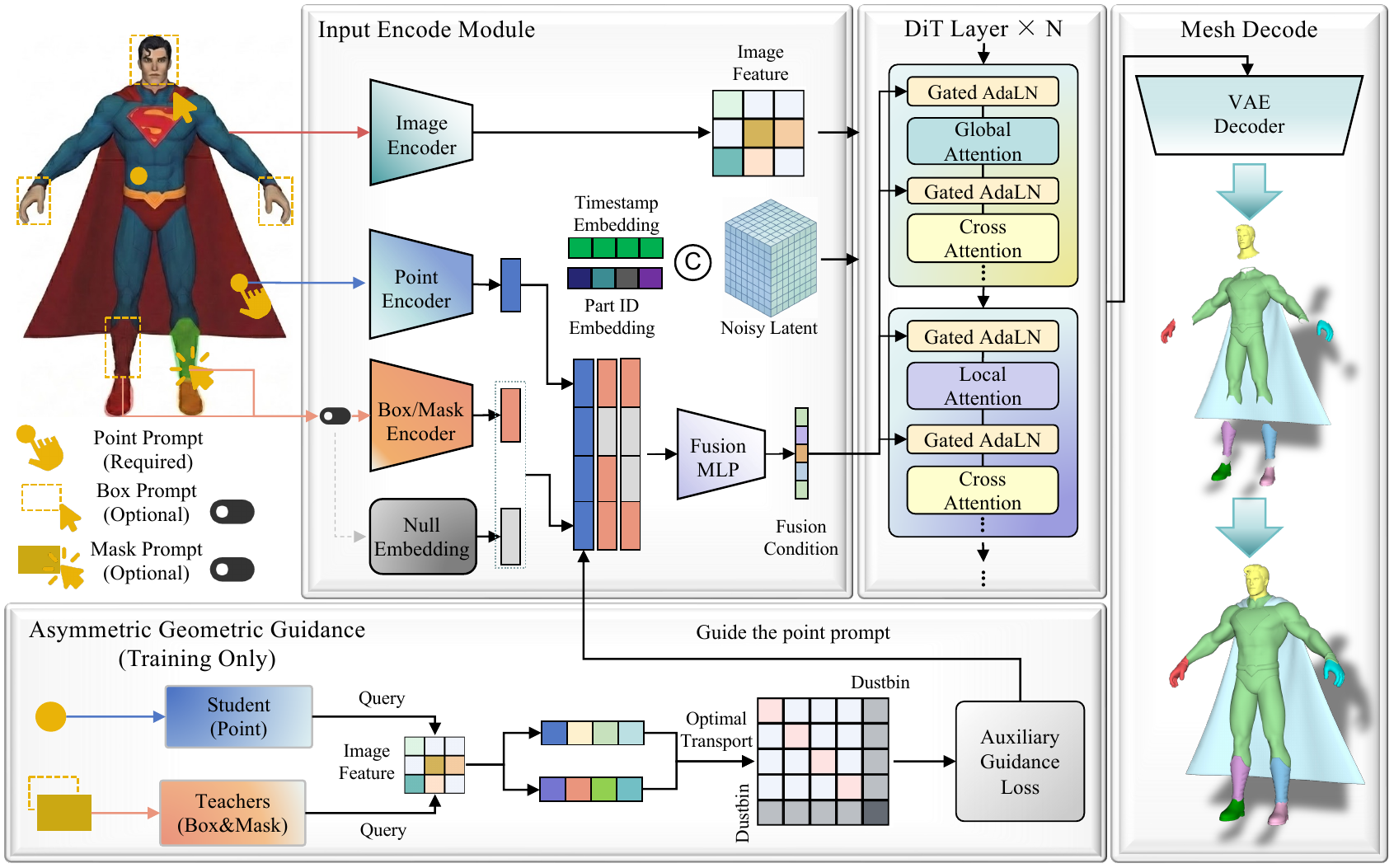}
  \caption{Overview of the \modelname framework. \modelname supports point, box, and mask prompts, with points as the minimal input. The noisy latent, part ID, and timestep are jointly conditioned on encoded prompts, where learnable null embeddings replace missing inputs. The fused condition modulates each DiT layer for part generation via gated adaptive layer normalization. During training, a multi-granularity geometric guidance branch aligns weak and strong priors via optimal transport.
  }
  \label{fig:Overview}
  
\end{figure*}

\section{Methodology}
Our method presents a unified multi-granularity framework for controllable image-to-3D part generation, accommodating diverse geometric user inputs, such as point prompts, 2D bounding boxes, and 2D masks, within a single model. Built upon the DiT architecture, \modelname leverages a Gated Adaptive Layer Normalization (Gated AdaLN) mechanism to inject these geometric conditions, enabling unified modeling of heterogeneous inputs with varying precision. To further support this versatility, we incorporate a geometric guidance mechanism to maintain robustness under weak supervision. The overall architecture is illustrated in Fig.~\ref{fig:Overview}.

\subsection{Gated Adaptive Modulation}

\textbf{Prompt Definition and Encoding.} 
We consider multiple geometric prompt types for part-level structure, including point, box, and mask prompts. For each part $i$, we denote their embeddings as $\mathbf{e}_i^{\mathrm{point}}, \mathbf{e}_i^{\mathrm{box}}, \mathbf{e}_i^{\mathrm{mask}} \in \mathbb{R}^{D}$, all produced by lightweight modality-specific encoders. Point prompts are encoded via Gaussian Fourier features followed by an MLP to capture continuous 2D locations. Box prompts encode top-left and bottom-right corners separately with a learnable corner-type embedding inspired by SAM~\cite{kirillov2023segment}. Mask prompts are processed using a small CNN with global average pooling to obtain compact representations. These encoders provide expressive, lightweight representations.

To train the model with different prompt combinations and improve robustness to missing inputs, we employ a probabilistic masking strategy. Although point prompts serve as the mandatory minimal input for users during inference, as shown in Fig.~\ref{fig:Overview}, we apply random masking to all modalities, including points, during training.
Specifically, we simulate missing geometric prompts by replacing them with learnable null embeddings. For each part $i$ and modality $k \in \{\text{point}, \text{box}, \text{mask}\}$, we use a binary drop mask $m_i^k \in \{0,1\}$ with probability $\Pr(m_i^k = 0) = p_\text{drop}$, where $p_\text{drop}$ is the manually set probability of masking a given modality. The effective prompt embedding is then defined as:
\begin{equation}
\tilde{\mathbf{e}}_i^k = m_i^k \mathbf{e}_i^k + (1-m_i^k) \mathbf{e}_{\text{null}}^k,
\end{equation}
so that any dropped or absent prompt is replaced by its learnable null embedding $\mathbf{e}_{\text{null}}^k \in \mathbb{R}^D$.

The resulting embeddings for each part are aggregated and projected through a learnable fusion MLP $\psi(\cdot)$ to yield a unified part-level conditioning vector:
\begin{equation}
\mathbf{c}_i = \psi\Big([\tilde{\mathbf{e}}_i^{\text{point}} \parallel \tilde{\mathbf{e}}_i^{\text{box}} \parallel \tilde{\mathbf{e}}_i^{\text{mask}}]\Big),
\end{equation}
where $[\cdot \parallel \cdot]$ denotes feature concatenation. This mechanism produces a conditioning vector $\mathbf{c}_i$ of fixed dimensionality for any input prompt combination.

\textbf{Multi-Prompt Conditional Modulation.}
Given the fused part condition vector $\mathbf{c}$, we inject geometric control into the DiT via Gated AdaLN. This module dynamically modulates the transformer hidden states, concatenated with the timestep $t$ and learnable part identity, by predicting scale and shift parameters based on $\mathbf{c}$.

A lightweight modulation network $\Phi(\cdot)$ maps $\mathbf{c}$ to adaptive scale, shift, and gating parameters $(\boldsymbol{\gamma}, \boldsymbol{\beta}, \mathbf{g}) = \Phi(\mathbf{c})$, which are applied to the hidden features $\mathbf{H}$ as
\begin{equation}
\mathbf{H}' = \mathbf{H} + \mathbf{g} \odot f\!\left((1+\boldsymbol{\gamma}) \odot \mathrm{LN}(\mathbf{H}) + \boldsymbol{\beta}\right),
\end{equation}
where $\mathrm{LN}(\cdot)$ denotes layer normalization and $f(\cdot)$ is a small feed-forward projection.
This gated formulation enables adaptive regulation of geometric conditions at each layer.
Complementing the geometric prompts, image features from DINOv2~\cite{Oquab2023DINOv2LR} are injected via cross-attention. Cross-attention provides global context, while Gated AdaLN controls part-level geometry.

\textbf{Part-Aware Attention.} Following the design of PartCrafter~\cite{lin2025partcrafter}, we adopt a part-aware attention mechanism in the DiT to model both global inter-part interactions and local intra-part geometry. In our setup, we treat individual parts as distinct samples. A batch contains $B$ parts, each represented by $L$ latent tokens, collectively forming a hidden state tensor $\mathbf{H} \in \mathbb{R}^{B \times L \times D}$.

\subsection{Asymmetric Geometric Guidance}
To enrich sparse point prompts with spatial semantics, we propose asymmetric alignment: strong priors (boxes/masks) supervise weak point representations via differentiable optimal transport.

\textbf{Asymmetric Visual Bridge.}
We project student point prompts and teacher box/mask prompts into a shared visual space using a Visual Bridge $\mathcal{B}(\cdot)$ with cross-attention over image features $\mathbf{I}_{\mathrm{enc}}$. To prevent the alignment loss from updating the teacher prompt embeddings, we apply stop-gradient to the box/mask embeddings before the Visual Bridge:
\begin{equation*} 
\mathbf{f}_i^{\mathrm{stu}} = \mathcal{B}(\mathbf{e}_i^{\mathrm{point}}, \mathbf{I}_{\mathrm{enc}}), \quad 
\mathbf{f}_i^{\mathrm{tea}} = \mathcal{B}(\mathrm{sg}(\mathbf{e}_i^{\mathrm{box/mask}}), \mathbf{I}_{\mathrm{enc}}). 
\end{equation*}

\textbf{Similarity-based Partial Optimal Transport.}  
We align student and teacher features via differentiable partial optimal transport~\cite{sarlin2020superglue}, which introduces global competition to resolve ambiguous matches. The affinity matrix is defined as:
\begin{equation}
\mathbf{S}_{ij} =
\tau \Big(
\alpha \langle \mathbf{f}_i^{\mathrm{stu}}, \mathbf{f}_j^{\mathrm{tea_{box}}} \rangle
+ (1-\alpha) \langle \mathbf{f}_i^{\mathrm{stu}}, \mathbf{f}_j^{\mathrm{tea_{mask}}} \rangle
\Big),
\end{equation}
where $\tau$ is a learnable temperature and $\alpha$ balances the two teacher sources.  

To account for missing conditions (e.g., classifier-free guidance~\cite{ho2022classifierfreediffusionguidance}), we introduce a learnable dustbin score $\phi$ and construct an augmented matrix $\bar{\mathbf{S}} \in \mathbb{R}^{(N+1)\times(M+1)}$ by padding $\mathbf{S}$ with $\phi$. The optimal transport plan is computed via log-space Sinkhorn:
\begin{equation}
\mathbf{P}^* = \arg\max_{\mathbf{P}\in\mathcal{U}} \sum_{i,j} \bar{\mathbf{S}}_{ij} \mathbf{P}_{ij} + \varepsilon H(\mathbf{P}),
\end{equation}
where $\mathcal{U}$ denotes valid transport plans and $H(\mathbf{P})$ is the entropy regularizer.

This mechanism transfers strong geometric priors from boxes and masks to point representations, improving spatial semantics without requiring dense prompts at inference.

\textbf{Training Objective.}  
We train \modelname with a unified objective that combines conditional Rectified Flow and feature alignment supervision:
\begin{equation}
\mathcal{L}
=
\mathbb{E}_{\mathcal{Z}_0, \boldsymbol{\epsilon}, t} 
\Big[
\big\|
(\boldsymbol{\epsilon} - \mathcal{Z}_0) - \mathbf{v}_\theta(\mathcal{Z}_t, t, \mathbf{c})
\big\|_2^2
\Big]
-
\lambda_{\mathrm{aux}} \frac{1}{N} \sum_{i=1}^{N} \log(\mathbf{P}_{i,i}^* + \delta),
\end{equation}
where the first term corresponds to the conditional Rectified Flow objective, encouraging accurate velocity prediction in the latent space, and the second term enforces index-wise alignment between student and teacher features via diagonal transport probabilities. Here, $N$ denotes the number of student point representations in the alignment term, $\delta$ is a small constant for numerical stability, and $\lambda_{\mathrm{aux}}$ controls the strength of the alignment supervision.

\textbf{Inference.} During inference, we support flexible geometric conditions ranging from sparse points to dense masks. Point prompts are always required as the base condition, while higher-level cues are incorporated hierarchically for flexible control. Specifically, masks can be paired with bounding boxes, and boxes can be combined with point prompts. This design enables precise control of part generation under varying levels of geometric guidance.

\section{Implementation Details and Metrics}
\label{sec:implementation}

\subsection{Implementation Details}  
\label{Implementation_Details}

We use a pretrained 3D VAE from TripoSG~\cite{li2025triposg} to obtain part latents. Each part is encoded into latent tokens following 3DShape2VecSet~\cite{zhang20233dshape2vecset}.
Following PartCrafter~\cite{lin2025partcrafter}, we initialize the generative backbone with pretrained TripoSG~\cite{li2025triposg} weights and fine-tune it using the following two-stage training strategy.
We train \modelname on a composite dataset derived from Objaverse~\cite{deitke2023objaverse} and PartVerseXL~\cite{ding2025fullpart}.
Stage 1 targets objects with 1--8 parts, each encoded with 512 tokens. To ensure robustness against suboptimal placement, we randomly sample points within masks and apply a 0.5 dropout rate to both boxes and masks. Stage 2 scales to 2--16 parts using 1024 tokens per part to handle varying granularities. In this stage, we set the dropout probability to 0.1 for points and 0.7 for masks, similar to~\cite{qu2025vincie,huang2023composer}. For bounding boxes, we implement a mixed dropping strategy with a probability of 0.25 for full dropout, and 0.25 for partial masking applied to half of the parts. Training is conducted on 8 NVIDIA H20 GPUs with a batch size of 256, using learning rates of $1\times10^{-4}$ and $5\times10^{-5}$ for the two stages, respectively.

\subsection{Datasets and Metrics}
\label{sec:datasets_metrics}
To assess generalization, we conduct evaluations on a held-out test set from PartVerseXL~\cite{ding2025fullpart} and PartObjaverse-Tiny~\cite{yang2024sampart3d}, ensuring strict disjointness from the training set. For each dataset, we select the view that maximizes the product of visible pixel count and number of visible parts, ensuring sufficient part coverage.

\textbf{Whole Mesh Evaluation Metrics.} We first assemble the generated parts into a complete mesh and evaluate overall geometric quality using Chamfer Distance (CD) and F-Score, where F-Score is computed with a distance threshold of 0.1. Semantic consistency between the generated mesh and the input image is measured using ULIP~\cite{Xue_2023_CVPR}.

\textbf{Part-level Evaluation Metrics.} A primary challenge in evaluating parts is that the number of generated components often differs from the ground truth. To ensure a fair comparison, we first filter out ground truth parts that are fully occluded in the input image. We apply CD-based Hungarian matching between the remaining ground-truth parts and generated parts to obtain one-to-one correspondences. We compute part-level metrics only over matched pairs; unmatched ground-truth and generated parts are excluded without additional penalties. Based on these matched pairs, we report Part CD to measure geometric deviation. Furthermore, we evaluate spatial alignment using part-level bounding-box IoU (IoU$_B$) for coarse extent, and surface-based voxel IoU (IoU$_V$) for fine-grained overlap. To handle non-watertight geometry, we define occupancy based on the distance field rather than interior filling. Specifically, we use a voxel grid with a pitch of $h_{\mathrm{vox}} = 0.03$, where a voxel is considered occupied if its distance to the nearest mesh surface is less than or equal to $h_{\mathrm{vox}}$. Unless otherwise stated, benchmark box inputs are derived from ground-truth masks, and points are sampled from the corresponding mask regions.

Detailed dataset construction and benchmark preparation are provided in \underline{\textbf{App.~\ref{Details_for_Datasets_and_Metrics}}}.

\section{Experiments} 
\label{sec:experiments}
\begin{table*}[t]
    \centering
    \scriptsize
    \setlength{\tabcolsep}{2pt}
    \caption{Part-level structure evaluation. Cond. denotes input conditions; PartPacker requires none. For whole-mesh baselines, we use SAM3D to lift multi-view 2D masks to 3D segmentations. Part IoU$_V$ and Part IoU$_B$ denote voxel IoU and 3D bounding box IoU. \textbf{Bold} and \underline{underlined} indicate the best and second-best results, respectively.}
    
    \begin{tabularx}{\textwidth}{lccYYYYYY}
        \toprule
        \multirow{2.3}{*}{Method} & \multirow{2.3}{*}{Reference} & \multirow{2.3}{*}{Cond.} & \multicolumn{3}{c}{\textbf{PartVerseXL}} & \multicolumn{3}{c}{\textbf{PartObjaverse-Tiny}} \\
        \cmidrule(lr){4-6} \cmidrule(lr){7-9}
         & & & Part CD $\downarrow$ & Part IoU$_V$ $\uparrow$ & Part IoU$_B$ $\uparrow$ & Part CD $\downarrow$ & Part IoU$_V$ $\uparrow$ & Part IoU$_B$ $\uparrow$ \\
        \midrule
        TRELLIS + SAM3D & CVPR 25 & MV Mask & 0.3471 & 0.1885 & 0.2173 & 0.3916 & 0.1226 & 0.2031 \\
        
        TripoSG + SAM3D & TPAMI 25 & MV Mask & 0.3523 & 0.1727 & 0.1896 & 0.4235 & 0.1250 & 0.2078 \\
        \midrule
        PartCrafter   & NeurIPS 25 & Number & 0.5235 & 0.0907 & 0.1257 & 0.5193 & 0.0802 & 0.1455 \\
        
        PartPacker    & NeurIPS 25 & - & 0.3933 & 0.1717 & 0.2210 & 0.3932 & \textbf{0.1588} & 0.2327 \\
        
        OmniPart      & SIGGRAPH Asia 25 & Mask & 0.3283 & 0.1961 & \underline{0.2831} & 0.3681 & 0.1568 & 0.2649 \\
        \midrule
        \rowcolor{linecolor!40}\modelname(\textbf{ours}) & - & Mask & \textbf{0.2938} & \underline{0.2049} & \textbf{0.2840} & \textbf{0.3560} & \underline{0.1571} & \textbf{0.2778} \\
        
        \rowcolor{linecolor!40}\modelname(\textbf{ours}) & - & Box & \underline{0.3042} & \textbf{0.2078} & 0.2810 & \underline{0.3675} & 0.1503 & \underline{0.2693} \\
        
        \rowcolor{linecolor!40}\modelname(\textbf{ours}) & - & Point & 0.3334 & 0.1902 & 0.2506 & 0.3802 & 0.1411 & 0.2456 \\
        \bottomrule
    \end{tabularx}
    \label{tab:part_structure}
    
\end{table*}

\begin{table*}[t]
    \centering
    \scriptsize
    \caption{Whole-mesh quality evaluation. TRELLIS and TripoSG are full-mesh generation methods, while FullPart overlaps with the PartVerseXL training set; these entries are grayed out and reported for reference only, and are excluded from direct comparison. \textbf{Bold} and \underline{underlined} indicate the best and second-best results.}
    
    \begin{tabularx}{\textwidth}{lcYYYYYY}
        \toprule
        \multirow{2.3}{*}{Method} & \multirow{2.3}{*}{Reference} & \multicolumn{3}{c}{\textbf{PartVerseXL}} & \multicolumn{3}{c}{\textbf{PartObjaverse-Tiny}} \\
        \cmidrule(lr){3-5} \cmidrule(lr){6-8}
         & & Mean CD $\downarrow$ & F-Score $\uparrow$ & ULIP $\uparrow$ & Mean CD $\downarrow$ & F-Score $\uparrow$ & ULIP $\uparrow$ \\
        \midrule
        
        \gr{TRELLIS} & \gr{CVPR 25} & \gr{0.2125} & \gr{0.7514} & \gr{0.2370} & \gr{0.2038} & \gr{0.7478} & \gr{0.2434} \\
        
        \gr{TripoSG} & \gr{TPAMI 25} & \gr{0.1227} & \gr{0.8176} & \gr{0.2331} & \gr{0.1504} & \gr{0.7811} & \gr{0.2410} \\
        \midrule
        
        PartCrafter & NeurIPS 25 & 0.1666 & 0.7346 & 0.1950 & 0.1731 & 0.7280 & 0.2136 \\
        
        PartPacker  & NeurIPS 25 & \underline{0.1489} & \textbf{0.7899} & \textbf{0.2266} & 0.1612 & \textbf{0.7682} & \textbf{0.2397}\\
        
        OmniPart    & SIGGRAPH Asia 25 & 0.1547 & 0.7636 & 0.2179 & \textbf{0.1585} & 0.7501 & 0.2335 \\
        
        FullPart    & ICLR 26 & - & - & - & 0.1625 & \underline{0.7577} & 0.1775 \\
        \midrule
        \rowcolor{linecolor!40}\modelname(\textbf{ours}) & - & \textbf{0.1418} & \underline{0.7877} & \underline{0.2206} & \underline{0.1587} & 0.7510 & \underline{0.2350} \\
        \bottomrule
    \end{tabularx}
    \label{tab:mesh_quality}
    
\end{table*}

\subsection{Main Results}

We compare \modelname against six state-of-the-art methods, including two holistic mesh generation models~\cite{xiang2025structured, li2025triposg} and four native part-based generation approaches~\cite{lin2025partcrafter, tang2025efficient, yang2025omnipart, ding2025fullpart}.

\graphicspath{{figures/}} 
\begin{figure*}[ht]
  \centering
  \includegraphics[width=\linewidth]{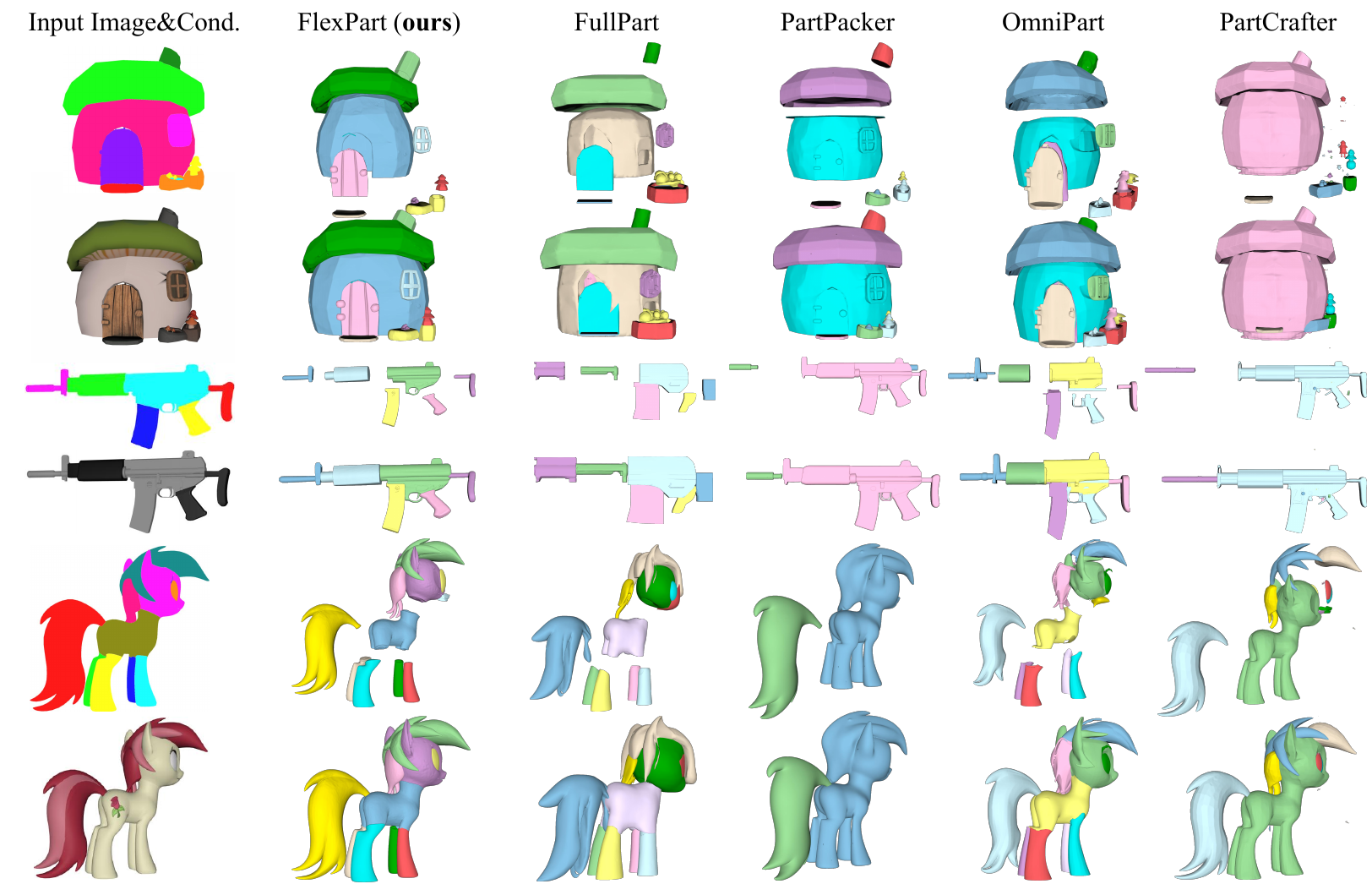}
  \caption {Qualitative results of different methods. The upper row shows exploded views for inspecting individual parts, and the lower row shows the assembled meshes. Our method uses masks as input, while other methods are evaluated with their respective input conditions.}
  \label{fig:qualitative_results}
  
\end{figure*}

\textbf{Part-level Mesh Quality.}
We first evaluate the geometric fidelity of individual parts (Tab.~\ref{tab:part_structure}) to assess controllable part generation. 
For comparison with whole-mesh generation methods, we employ SAM3D~\cite{yang2023sam3d} to project multi-view masks onto their generated meshes for part segmentation. This post-hoc lifting process often introduces geometric inconsistencies, significantly degrading part-level accuracy.
In contrast, \modelname directly infers compact part geometry. Under the same mask condition, our method outperforms OmniPart by a large margin, reducing Part CD on PartVerseXL by 10.5\%. This improvement is consistent with the benefits of direct part generation: OmniPart uses intermediate axis-aligned 3D boxes, which can enclose substantial empty space around non-axis-aligned shapes.

Crucially, \modelname maintains robust performance across varying input granularities. Relaxing the condition from masks to boxes yields stable quality, with Part CD degrading by merely 3.5\% on PartVerseXL. Even under these weaker priors, \modelname achieves better Part CD and Part IoU$_B$ than structurally unconstrained methods like PartPacker. FullPart is excluded here due to its reliance on 3D bounding boxes. A fairer evaluation under controlled settings is provided in Sec.~\ref{more_analysis}.

\textbf{Whole Mesh Quality.} 
Complementing our superior part-level performance, we further assess the global quality of the assembled meshes. As shown in Tab.~\ref{tab:mesh_quality}, \modelname achieves strong and consistent mesh-level quality across both benchmarks. On PartVerseXL, we achieve a Mean CD of 0.1418, surpassing OmniPart by 8\%, indicating that our improvements in local part precision translate into better global geometric accuracy. On PartObjaverse-Tiny, while some specialized methods achieve marginally better results, \modelname remains highly competitive. These results show that \modelname supports flexible part control while maintaining competitive whole-mesh quality.

\textbf{Qualitative Results.}
Fig.~\ref{fig:qualitative_results} presents qualitative comparisons across diverse categories. FullPart provides rough layout control via boxes but often yields inaccurate shapes. PartPacker achieves decent global geometry but lacks explicit control to match user-specified structures. OmniPart tends to distort meshes, deviating noticeably from inputs, while PartCrafter, limited to part-count conditioning, struggles to align with intended divisions. In contrast, \modelname generates distinct and semantically meaningful parts, ensuring visual realism and structural coherence suitable for downstream editing.

\subsection{Analysis and Ablation Study}
\label{more_analysis}
To investigate the architecture design, we conduct ablation studies on PartVerseXL unless otherwise specified. Default settings are marked in \colorbox{linecolor!40}{gray}.

\begin{figure}[t]
  \centering
  \includegraphics[width=\linewidth]{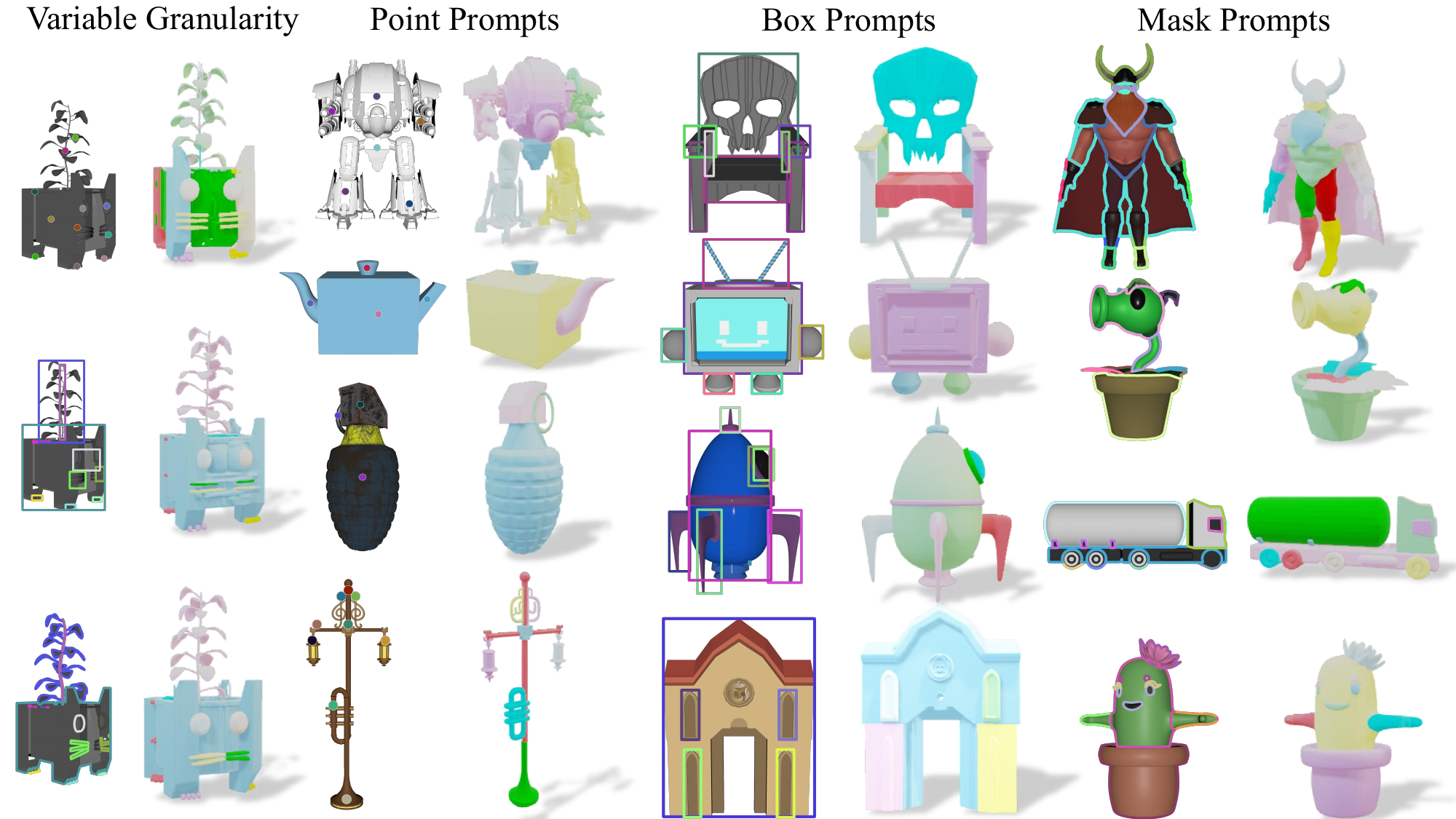}
  \caption {Qualitative generation results of our method under varying geometric prompts.}
  \label{fig:comparison_granularity}
  
\end{figure}

\textbf{Effect of Multi-granularity Synergy.} Tab.~\ref{tab:unified_ablation_synergy} demonstrates that our framework achieves deep cross-modal synergy instead of a naive combination of conditions. Specifically, stronger geometric priors effectively guide weaker ones during training. Our joint model improves \textit{Point only} inference by 7.9\% in IoU$_V$ and 6.1\% in CD because it internalizes precise boundary constraints from dense inputs. Furthermore, joint training consistently outperforms single-modality models across all evaluation settings. For example, under mask conditioning, the unified model reduces Part CD from 0.3224 for the mask-only training baseline to 0.2938. These results confirm that \modelname learns integrated semantics in a shared space. This formulation provides a versatile solution that maintains high geometric fidelity across diverse interaction modes. Fig.~\ref{fig:comparison_granularity} illustrates robust, semantically consistent results across prompt granularities.

\begin{table}[t]
    \centering
    \scriptsize 
    \setlength{\tabcolsep}{8pt}
    \caption{Ablation on multi-granularity prompt synergy. Eval Cond. denotes the evaluation condition, while Train Setting lists the prompt modalities used to train each model. \textit{Point only}, \textit{Box only}, and \textit{Mask only} denote single-modality training baselines. We compare whole-mesh and part-level fidelity within each evaluation condition.}
    \begin{tabular}{ll ccc ccc} 
        \toprule
        \multirow{2.3}{*}{\textbf{Eval Cond.}} & \multirow{2.3}{*}{\textbf{Train Setting}} & \multicolumn{3}{c}{\textbf{Whole Mesh}} & \multicolumn{3}{c}{\textbf{Part-level}} \\
        \cmidrule(lr){3-5} \cmidrule(lr){6-8}
         & & CD $\downarrow$ & F-Score $\uparrow$ & ULIP $\uparrow$ & CD $\downarrow$ & IoU$_V$ $\uparrow$ & IoU$_B$ $\uparrow$ \\
        \midrule
        \multirow{3}{*}{\textbf{Point}} 
         & Point only & 0.1623 & 0.7368 & 0.2014 & 0.3550 & 0.1763 & 0.2320 \\
         & Point + Mask & 0.1587 & 0.7505 & 0.2112 & 0.3421 & 0.1816 & 0.2396 \\
         & \cellcolor{linecolor!40}Point + Box + Mask & \cellcolor{linecolor!40}\textbf{0.1525} & \cellcolor{linecolor!40}\textbf{0.7689} & \cellcolor{linecolor!40}\textbf{0.2171} & \cellcolor{linecolor!40}\textbf{0.3334} & \cellcolor{linecolor!40}\textbf{0.1902} & \cellcolor{linecolor!40}\textbf{0.2506} \\
        \midrule
        \multirow{4}{*}{\textbf{Box}} 
         & Box only & 0.1489 & 0.7756 & 0.2171 & 0.3155 & 0.2011 & 0.2756 \\
         & Point + Box & 0.1458 & 0.7812 & 0.2201 & 0.3102 & 0.2023 & 0.2763 \\
         & Box + Mask & \textbf{0.1434} & \textbf{0.7834} & 0.2209 & 0.3094 & \textbf{0.2089} & 0.2795 \\
         & \cellcolor{linecolor!40}Point + Box + Mask & \cellcolor{linecolor!40}0.1447 & \cellcolor{linecolor!40}0.7828 & \cellcolor{linecolor!40}\textbf{0.2219} & \cellcolor{linecolor!40}\textbf{0.3042} & \cellcolor{linecolor!40}0.2078 & \cellcolor{linecolor!40}\textbf{0.2810} \\
        \midrule
        \multirow{4}{*}{\textbf{Mask}} 
         & Mask only & 0.1541 & 0.7564 & 0.2056 & 0.3224 & 0.1905 & 0.2688 \\
         & Point + Mask & 0.1501 & 0.7712 & 0.2169 & 0.3202 & 0.1921 & 0.2712 \\
         & Box + Mask & 0.1450 & 0.7848 & 0.2195 & 0.3076 & 0.1998 & 0.2822 \\
         & \cellcolor{linecolor!40}Point + Box + Mask & \cellcolor{linecolor!40}\textbf{0.1418} & \cellcolor{linecolor!40}\textbf{0.7877} & \cellcolor{linecolor!40}\textbf{0.2206} & \cellcolor{linecolor!40}\textbf{0.2938} & \cellcolor{linecolor!40}\textbf{0.2049} & \cellcolor{linecolor!40}\textbf{0.2840} \\
        \bottomrule
    \end{tabular}
    \label{tab:unified_ablation_synergy}
    
\end{table}

\graphicspath{{figures/}} 

\Needspace{16\baselineskip}
\begin{wraptable}{r}{0.46\textwidth}
    \centering
    \scriptsize
    \setlength{\tabcolsep}{3.5pt}
    \captionsetup{font=small,skip=5pt,justification=RaggedRight,singlelinecheck=false}
    \caption{Ablation on Asymmetric Geometric Gu\-idance under point and mask prompts.}
    \begin{tabular*}{\linewidth}{@{\extracolsep{\fill}}lcccc}
        \toprule
        \multirow{2.3}{*}{\makecell{Cond.}} & \multirow{2.3}{*}{Guid.} & \multicolumn{3}{c}{\textbf{Part-level}} \\
        \cmidrule(lr){3-5}
         & & CD $\downarrow$ & IoU$_V$ $\uparrow$ & IoU$_B$ $\uparrow$ \\
        \midrule
        \multirow{2}{*}{Points} & - & 0.3406 & 0.1811 & 0.2323 \\
         & \cellcolor{linecolor!40}\checkmark & \cellcolor{linecolor!40}\textbf{0.3334} & \cellcolor{linecolor!40}\textbf{0.1902} & \cellcolor{linecolor!40}\textbf{0.2506} \\
        \midrule
        \multirow{2}{*}{Masks} & - & 0.3011 & \textbf{0.2078} & 0.2813 \\
         & \cellcolor{linecolor!40}\checkmark & \cellcolor{linecolor!40}\textbf{0.2938} & \cellcolor{linecolor!40}0.2049 & \cellcolor{linecolor!40}\textbf{0.2840} \\
        \bottomrule
    \end{tabular*}
    \label{tab:ablation_guidance}
\end{wraptable}
\textbf{Effect of Asymmetric Geometric Guidance.}\newline
Building upon the observed multi-granularity synergy, we evaluate the proposed Asymmetric Geometric Guidance by ablating the auxiliary alignment loss. As shown in Tab.~\ref{tab:ablation_guidance}, it significantly improves performance under sparse point prompts, yielding gains of 7.9\% in Part IoU$_B$ and 5.0\% in Part IoU$_V$ over the baseline. This indicates that OT-based alignment effectively transfers structural priors from dense annotations to regularize learning from sparse inputs. Performance under mask conditions remains comparable, suggesting positive cross-granularity transfer without conflict. The guidance operates only during training, with no inference overhead.

\Needspace{16\baselineskip}
\begin{wraptable}{r}{0.62\textwidth}
    \centering
    \scriptsize
    \setlength{\tabcolsep}{2.5pt}
    \captionsetup{font=small,skip=5pt,justification=raggedright,singlelinecheck=false}
    \caption{Comparison with FullPart under 3D box conditioning on the PartObjaverse-Tiny benchmark.}
    \label{tab:stronger_cond}
    \begin{tabular*}{\linewidth}{@{\extracolsep{\fill}}lccccccc}
        \toprule
        \multirow{2.3}{*}{Method} & \multirow{2.3}{*}{Cond.} & \multicolumn{3}{c}{\textbf{Whole Mesh}} & \multicolumn{3}{c}{\textbf{Part-level}} \\
        \cmidrule(lr){3-5} \cmidrule(lr){6-8}
         & & CD$\downarrow$ & F-Score$\uparrow$ & ULIP$\uparrow$ & CD$\downarrow$ & IoU$_V$ $\uparrow$ & IoU$_B$ $\uparrow$ \\
        \midrule
        FullPart & 3D Box & 0.1625 & 0.7577 & 0.1775 & 0.2488 & 0.2064 & \textbf{0.4370} \\
        \rowcolor{linecolor!40}Ours & 3D Box & \textbf{0.1385} & \textbf{0.8158} & \textbf{0.2209} & \textbf{0.2356} & \textbf{0.2653} & 0.4027 \\
        \bottomrule
    \end{tabular*}
\end{wraptable}
\textbf{Analysis of Stronger 3D Conditions.}
Although our primary focus is on generating 3D parts from lightweight 2D inputs, we also evaluate \modelname under explicit 3D constraints. For a fair comparison with FullPart, which uses 3D bounding boxes, we retrain a variant conditioned solely on 3D boxes and evaluate it on PartObjaverse-Tiny. As shown in Tab.~\ref{tab:stronger_cond}, our method outperforms FullPart on most metrics under the same supervision. This result demonstrates that Gated AdaLN handles both sparse 2D inputs and explicit 3D constraints.

\textbf{Analysis of SAM-based Alternatives.}
A seemingly straightforward alternative is to convert sparse prompts into dense masks using 2D segmentation models (e.g., SAM) before 3D generation. However, such cascaded pipelines are inherently limited: lacking 3D part-level semantics, SAM often produces localized fragments rather than complete functional parts, leading to accumulated errors and degraded geometry. In contrast, FlexPart directly models heterogeneous conditions end-to-end, avoiding intermediate information loss and enabling consistent part-level generation from sparse prompts. Detailed analyses are provided in \underline{\textbf{App.~\ref{Appendix:SAM_Analysis}}}.

\textbf{Robustness to Input Variations.} Perturbing the position and size of 2D boxes by up to 20\% increases Part CD from 0.3042 to 0.3171, while Part IoU$_V$ changes from 0.2078 to 0.2032. Providing boxes for half of the parts and points for the remainder yields Part CD of 0.3287. Mixing masks for half of the parts with boxes for the rest achieves Part CD of 0.3190, despite this combination being unseen during training. These results support generation with noisy and heterogeneous prompts. A complementary study evaluates joint point-and-box inputs from three annotators on 50 PartObjaverse-Tiny objects. Full protocols and results, including count-only inputs, are provided in \underline{\textbf{App.~\ref{User_Input_Variations} and \ref{app:human_prompts}}}.

\textbf{Sensitivity of Point Prompts.}
We evaluate four point-placement strategies on PartVerseXL: \textit{Center} (the part-mask centroid), \textit{Edge} (near the part boundary), \textit{Offset} (a perturbed centroid), and \textit{Cluster} (the patch center closest to the object-level geometric center). Under point-only input, switching from Center to Edge increases Part CD from 0.3302 to 0.4190 and reduces Part IoU$_V$ from 0.1905 to 0.1342. With additional box or mask prompts, Part CD varies across the four strategies within 0.3036--0.3178 and 0.2995--0.3087, respectively. Stronger geometric prompts therefore reduce sensitivity to point placement. Experimental details and quantitative and qualitative results appear in \underline{\textbf{App.~\ref{Appendix:point_sensitivity}}}.

\section{Conclusion}
\label{sec:conclusion}
We introduced \modelname, a unified framework for controllable image-to-3D part generation with geometric prompts at varying granularities. Gated adaptive modulation integrates heterogeneous prompts within a single model, while Asymmetric Geometric Guidance uses strong spatial priors to improve sparse point representations during training. Experiments demonstrate improved part-level geometric quality and competitive whole-mesh fidelity across prompt conditions. These results support joint multi-granularity learning as a practical approach to flexible 3D part generation.

\textbf{Limitations.} Occlusion handling remains challenging. Single-view 2D conditions provide limited depth cues, and the model may place occluded parts incorrectly in visible regions. Improving spatial reasoning under occlusion is a direction for future work.

\FloatBarrier
\renewcommand{\acksection}{\section*{Acknowledgments}}
\begin{ack}
We thank Chongqing Afari Technology Co., Ltd. for providing computational resources.

This work was supported by The Explorers Program of Shanghai (Basic Research Funding) (26TS1404200); the Fundamental and Interdisciplinary Disciplines Breakthrough Plan of the Ministry of Education of China (JYB2025XDXM312); and Tongji University's ``Medicine+X'' Cultivation Program (2026-0648-YB-05).
\end{ack}

{\small
\setlength{\bibsep}{2pt plus 1pt minus 1pt}
\bibliographystyle{plain}
\bibliography{references}
}

\clearpage
\newpage
\appendix
\renewcommand{\topfraction}{0.9}
\renewcommand{\bottomfraction}{0.8}
\renewcommand{\textfraction}{0.08}
\renewcommand{\floatpagefraction}{0.8}
\setcounter{topnumber}{3}
\setcounter{bottomnumber}{2}
\setcounter{totalnumber}{5}
\clubpenalty=10000
\widowpenalty=10000

\section{Dataset Preparation and Benchmark Details} 
\label{Details_for_Datasets_and_Metrics}

\textbf{Data Preparation.}
We construct the training set from subsets of Objaverse~\cite{deitke2023objaverse} and PartVerseXL~\cite{ding2025fullpart}, totaling approximately 250K 3D models. Due to batch-wise part-packing during training, about 180K models are effectively used. The Objaverse subset provides heterogeneous annotations, where some models contain only complete object meshes while others include part-level meshes. PartVerseXL complements this with densely and carefully annotated part-level meshes. Since most Objaverse meshes are non-watertight, we convert all models into watertight representations, following prior work~\cite{chen2025dora,li2025triposg,yang2024hunyuan3d,zhao2025hunyuan3d}, showing that watertight geometry leads to more stable training and improved generation quality. For each mesh, multiple viewpoints are rendered, and the view with the largest visible foreground area is selected as the training image.

For the selected view, we generate 2D part visibility masks via ID-based rendering. Each part is rendered with a unique color, and depth buffering naturally resolves inter-part occlusions. The rendered image is decoded into a part ID map, from which binary visibility masks are extracted. Bounding boxes and point prompts are then derived directly from the corresponding masks, enabling unified geometric supervision at multiple granularities. To simulate user interactions, we sample 10 points within each mask and select the one nearest its centroid.

\textbf{Benchmarks.}
For evaluation, we construct benchmark subsets to test generalization. From PartVerseXL~\cite{ding2025fullpart}, which provides high-quality part annotations, we create a held-out test set with no overlap with our training data by randomly selecting 100 shapes with no more than 16 annotated parts, matching the maximum part count used during training. We also evaluate on PartObjaverse-Tiny~\cite{yang2024sampart3d}, a publicly available part-annotated dataset, including all shapes with up to 16 parts. Although some annotations in PartObjaverse-Tiny are less precise, it provides a complementary setting for assessing part-level generalization. For each dataset, we select the view that maximizes the product of visible pixel count and the number of visible parts, ensuring an informative view with sufficient part coverage. Before computing geometric metrics, we align each generated mesh to its ground-truth counterpart using rigid Iterative Closest Point.

\subsection{Evaluation Metric Definitions}
\label{app:metric_definitions}
Before ICP, each shape is normalized to fit within $[-1,1]^3$ so that its longest bounding-box side is 2, with no additional scaling during rigid alignment.

\textbf{Whole Mesh Evaluation Metrics.} Whole CD is the symmetric unsquared L2 Chamfer distance over 10K surface samples. For sampled point sets $X$ and $Y$,
\begin{equation}
\begin{aligned}
\operatorname{CD}(X,Y) ={}& \frac{1}{|X|}\sum_{x\in X}\min_{y\in Y}\|x-y\|_2
&+ \frac{1}{|Y|}\sum_{y\in Y}\min_{x\in X}\|y-x\|_2.
\end{aligned}
\end{equation}
 F-Score is the harmonic mean of precision and recall at distance threshold 0.1 (5\% of the normalized longest side). ULIP~\cite{Xue_2023_CVPR} measures semantic consistency with the input image. The same CD definition is used for matched part pairs.

\textbf{Part-level Evaluation Metrics.} A primary challenge in evaluating parts is that the number of generated components often differs from the ground truth. To ensure a fair comparison, we first filter out ground truth parts that are fully occluded in the input image. We apply CD-based Hungarian matching between the remaining ground-truth parts and generated parts to obtain one-to-one correspondences. We compute part-level metrics only over matched pairs; unmatched ground-truth and generated parts are excluded without additional penalties. Based on these matched pairs, we report Part CD to measure geometric deviation. Furthermore, we evaluate spatial alignment using part-level bounding-box IoU (IoU$_B$) for coarse extent, and surface-based voxel IoU (IoU$_V$) for fine-grained overlap. To handle non-watertight geometry, we define occupancy based on the distance field rather than interior filling. Specifically, we use a voxel grid with a pitch of $h_{\mathrm{vox}} = 0.03$, where a voxel is considered occupied if its distance to the nearest mesh surface is less than or equal to $h_{\mathrm{vox}}$. The normalized longest axis spans approximately 67 voxels at this pitch, which is fixed across all methods. Unless otherwise stated, benchmark boxes are derived from masks, and points are sampled from the corresponding mask regions.

\subsection{Baseline Protocol and Data Lineage}
\label{app:baseline_protocol}
All baselines use official checkpoints, without retraining or fine-tuning. TRELLIS and TripoSG receive an image, with multi-view SAM3D lifting for part evaluation; PartCrafter receives an image and GT part count; PartPacker receives only an image; OmniPart receives an image and GT-derived 2D masks; FullPart receives an image and GT 3D boxes. We re-evaluate every method using the same view selection, shape normalization, ICP alignment, and metrics. The main table is a system-level comparison with different inputs and training sets. Controlled comparisons use mask inputs for OmniPart, 3D boxes for FullPart, and part count for PartCrafter.

TRELLIS uses TRELLIS-500K; TripoSG uses curated Objaverse(-XL) and ShapeNet data. 
PartCrafter uses part-annotated Objaverse, ShapeNet, and ABO data; PartPacker uses filtered/extracted parts from TRELLIS-500K/Objaverse-XL; 
OmniPart uses self-constructed part-labeled data with a TRELLIS-based backbone. FullPart trains on PartVerseXL and is omitted on that benchmark. Our PartVerseXL evaluation set is strictly held out, whereas PartObjaverse-Tiny shares Objaverse asset lineage with several pretrained methods, including our TripoSG initialization.

\subsection{Alignment Implementation and Overhead}
\label{app:ot_implementation}
We use 50 log-domain Sinkhorn iterations with a learnable dustbin score and temperature. The box/mask fusion weight is $\alpha=0.5$, and the auxiliary alignment loss weight is $\lambda_{\mathrm{aux}}=1.0$. The alignment branch has 1.18M parameters (0.077\% of the model), and transport matrices are at most $33\times33$. An fp16 profile on one NVIDIA H20 GPU, with 8 parts and 1024 latent tokens per part, measures 8.9 ms for this branch versus 525 ms for the DiT forward pass (1.70\%). This measures forward-pass overhead in the profiled configuration, not total training throughput.


\section{Conditioning Route Analysis}
\label{app:conditioning}
We compare Gated AdaLN with injecting geometric prompt tokens into the pretrained image cross-attention pathway. The token-injection variant uses the full training schedule. Under mask conditioning, it performs worse than Gated AdaLN and the count-conditioned PartCrafter reference (Table~\ref{tab:conditioning_route}). A possible explanation is competition between prompt tokens and the 257 DINOv2 tokens within one attention softmax; the separate modulation route avoids this competition. This diagnostic supports our choice for this backbone, without establishing a general disadvantage of cross-attention conditioning.
\begin{table}[t]
\centering\small
\caption{Conditioning routes under mask input on PartVerseXL. PartCrafter is a reference using part count and no spatial prompts, not a matched-condition ablation.}
\label{tab:conditioning_route}
\begin{tabular}{lccc}
\toprule
Conditioning route & Part CD $\downarrow$ & IoU$_V$ $\uparrow$ & IoU$_B$ $\uparrow$ \\
\midrule
Gated AdaLN (ours) & 0.2938 & 0.2049 & 0.2840 \\
Prompt tokens in cross-attention & 0.6147 & 0.0818 & 0.1089 \\
PartCrafter (reference) & 0.5235 & 0.0907 & 0.1257 \\
\bottomrule
\end{tabular}
\vspace{-15pt}
\end{table}

\section{Analysis of Asymmetric Geometric Guidance}
\label{app:guidance_analysis}

\subsection{Effect of Stop-Gradient Design}

\begin{table}[t]
    \centering
    \scriptsize 
    \setlength{\tabcolsep}{8pt}
    \caption{Ablation on the Stop-Gradient Design in Geometric Guidance. "w/ Grad" allows the alignment loss to update the teacher prompt embeddings, while "wo/ Grad" (our default) stops gradients at the box/mask embeddings before the Visual Bridge.}
    
    \begin{tabular}{llccc}
        \toprule
        \textbf{Condition} & \textbf{Setting} & CD $\downarrow$ & F-Score $\uparrow$ & ULIP $\uparrow$ \\
        \midrule
        \multirow{2}{*}{Point} 
         & w/ Grad & \textbf{0.1518} & \textbf{0.7702} & 0.2166 \\
         & \cellcolor{linecolor!40}wo/ Grad & \cellcolor{linecolor!40}0.1525 & \cellcolor{linecolor!40}0.7689 & \cellcolor{linecolor!40}\textbf{0.2171} \\
        \midrule
        \multirow{2}{*}{Box} 
         & w/ Grad & 0.1531 & 0.7762 & 0.2198 \\
         & \cellcolor{linecolor!40}wo/ Grad & \cellcolor{linecolor!40}\textbf{0.1447} & \cellcolor{linecolor!40}\textbf{0.7828} & \cellcolor{linecolor!40}\textbf{0.2219} \\
        \midrule
        \multirow{2}{*}{Mask} 
         & w/ Grad & 0.1512 & 0.7761 & \textbf{0.2212} \\
         & \cellcolor{linecolor!40}wo/ Grad & \cellcolor{linecolor!40}\textbf{0.1418} & \cellcolor{linecolor!40}\textbf{0.7877} & \cellcolor{linecolor!40}0.2206 \\
        \bottomrule
    \end{tabular}
    \label{tab:stop_gradient_ablation}
    
\end{table}

We further analyze the effect of the stop-gradient design in geometric guidance. We find that it primarily affects whole-mesh reconstruction quality across different prompt granularities. As shown in Tab.~\ref{tab:stop_gradient_ablation}, enabling gradient flow (w/ Grad) slightly improves performance under sparse point prompts, reducing CD from 0.1525 to 0.1518. However, it degrades performance under stronger geometric conditions, including both box and mask prompts, where CD increases from 0.1447 to 0.1531 and from 0.1418 to 0.1512, respectively.

This behavior suggests that allowing gradients leads to overly symmetric optimization, where strong and weak prompts are pulled toward each other in representation space. Our goal is to guide weak point representations using strong geometric priors while preventing the alignment loss from updating the teacher prompt embeddings. We stop gradients at box/mask embeddings before the Visual Bridge.

\subsection{Partial OT versus Contrastive Alignment}
\label{app:ot_loss_comparison}

We evaluate whether partial optimal transport (OT) improves generation beyond a simpler contrastive alignment objective. We train an InfoNCE variant from the same initialization and with the same training budget, changing only the alignment loss. Table~\ref{tab:ot_infonce} reports the resulting part-level generation metrics on PartVerseXL. Partial OT achieves lower Part CD and higher voxel and bounding-box IoU under all three prompt conditions. In particular, under point prompts, Part CD decreases from 0.3499 to 0.3334. This experiment evaluates the alignment objective in end-to-end generation.

\begin{table}[t]
    \centering
    \small
    \caption{End-to-end comparison of alignment losses on PartVerseXL. Only the alignment loss is changed; initialization and training budget are held fixed. IoU$_V$ and IoU$_B$ denote part-level voxel and bounding-box IoU, respectively.}
    \label{tab:ot_infonce}
    \begin{tabular}{llccc}
        \toprule
        Condition & Alignment loss & Part CD $\downarrow$ & IoU$_V$ $\uparrow$ & IoU$_B$ $\uparrow$ \\
        \midrule
        Mask & Partial OT (ours) & \textbf{0.2938} & \textbf{0.2049} & \textbf{0.2840} \\
             & InfoNCE & 0.3053 & 0.1924 & 0.2673 \\
        \midrule
        Box & Partial OT (ours) & \textbf{0.3042} & \textbf{0.2078} & \textbf{0.2810} \\
            & InfoNCE & 0.3074 & 0.1874 & 0.2653 \\
        \midrule
        Point & Partial OT (ours) & \textbf{0.3334} & \textbf{0.1902} & \textbf{0.2506} \\
              & InfoNCE & 0.3499 & 0.1807 & 0.2346 \\
        \bottomrule
    \end{tabular}
    \vspace{-5pt}
\end{table}

\subsection{Global Assignment on Frozen Features}
\label{app:ot_assignment}

Known part correspondences provide diagonal supervision during training, but point features can still be ambiguous among similar strong priors. Paired L2 alignment treats each known pair independently, while InfoNCE normalizes similarities independently within each point row. Neither imposes a column marginal on the induced matching. Sinkhorn instead couples the assignments through both row and column marginals, introducing competition among points for the available priors.

To isolate this assignment effect, we freeze the trained model and use 199 PartObjaverse-Tiny objects containing 1,330 parts. Point, box, and mask prompts are derived from each ground-truth part as in training. We run only the prompt encoders, visual bridge, and DINOv2 feature extractor, without diffusion sampling or mesh decoding. We then apply row-wise similarity matching and Sinkhorn to the same similarity matrix. Top-1 matching measures correct point-to-prior matches, while many-to-one collapse measures duplicated prior selections by multiple points.

\begin{table}[t]
    \centering
    \small
    \caption{Assignment-operator diagnostic on the same frozen features. These are feature-matching results, rather than generation metrics or separate L2/InfoNCE training experiments.}
    \label{tab:ot_assignment}
    \begin{tabular}{lcc}
        \toprule
        Assignment operator & Top-1 matching $\uparrow$ & Many-to-one collapse $\downarrow$ \\
        \midrule
        Row-wise similarity matching & 0.803 & 0.133 \\
        Sinkhorn global assignment & \textbf{0.895} & \textbf{0.024} \\
        \bottomrule
    \end{tabular}
\end{table}

On normalized features, an InfoNCE similarity argmax and an L2 nearest-neighbor readout both reduce to row-wise similarity matching, although their training objectives differ. As shown in Table~\ref{tab:ot_assignment}, Sinkhorn improves Top-1 matching from 0.803 to 0.895 and reduces many-to-one collapse by approximately $5.5\times$. Ambiguity is particularly pronounced for symmetric parts: row-wise matching accuracy is 0.516 for symmetric pairs, compared with 0.931 for other pairs. These observations support global competition for resolving ambiguous feature correspondences.

\subsection{Dustbin for Unmatched Priors}
\label{app:ot_dustbin}

The dustbin addresses a separate issue: unmatched entries in batch-level alignment. In the same frozen-feature diagnostic, we append four box-prior columns from another object, none of which corresponds to a part of the current object. We compare strict OT with partial OT using the dustbin under this shared distractor setting. As shown in Table~\ref{tab:ot_dustbin}, partial OT improves Top-1 matching from 0.677 to 0.747. The dustbin supplies a channel for unmatched mass, relaxing the requirement to assign mass to irrelevant priors. Together with the preceding diagnostic, this separates the role of global competition among valid matches from the role of unmatched-mass handling.

\begin{table}[t]
    \centering
    \small
    \caption{Frozen-feature diagnostic with four cross-object box-prior distractors. Both variants receive the same augmented set of priors; the distractors have no valid correspondence in the current object.}
    \label{tab:ot_dustbin}
    \begin{tabular}{lc}
        \toprule
        Transport formulation & Top-1 matching $\uparrow$ \\
        \midrule
        Strict OT & 0.677 \\
        Partial OT with dustbin & \textbf{0.747} \\
        \bottomrule
    \end{tabular}
\end{table}

\section{Adaptability to Input Variations}
\label{User_Input_Variations}

\begin{table}[t]
    \centering
    \scriptsize 
    \caption{Analysis of the adaptability to user input variations.}
    \setlength{\tabcolsep}{10pt}
    \begin{tabular}{lccc}
        \toprule
        \multirow{2.3}{*}{Method} & \multicolumn{3}{c}{\textbf{Part-level}} \\
        \cmidrule(lr){2-4}
         & CD $\downarrow$ & IoU$_V$ $\uparrow$ & IoU$_B$ $\uparrow$ \\
        \midrule
        
        2D Box w/ Perturbation       & 0.3171 & 0.2032 & 0.2759 \\
        
        Half Box + Point             & 0.3287 & 0.1998 & 0.2609  \\

        Half Mask + Half Box         & 0.3190 & 0.2053 & 0.2706  \\

        Part Count Only              & 0.4786 & 0.1255 & 0.1608  \\

        \rowcolor{linecolor!40}\textbf{Ours (2D Box)}       & 0.3042 & 0.2078 & 0.2810 \\
        \bottomrule
    \end{tabular}
    \label{tab:input_analysis}
    
\end{table}

We evaluate \modelname under noisy, partial, and heterogeneous inputs through controlled experiments that approximate common user input variations. Drawn bounding boxes may be misaligned, loosely cover target parts, or specify only a subset of components.

\textbf{Perturbed bounding boxes.}
We simulate noisy user annotations by randomly offsetting the position and size of each input 2D bounding box by up to 20\%. This perturbation tests the model's tolerance to spatial inaccuracy while preserving approximate part localization.

\textbf{Partial box specification.}
To reflect scenarios where users focus on only a subset of parts, we provide bounding boxes for half of the parts and use point prompts for the remaining parts. This hybrid setting evaluates whether strong and weak geometric prompts can be jointly exploited within the unified framework.

\textbf{Mixed-condition specification.}
To further assess the effectiveness of our unified framework, we consider a mixed condition setting that is never observed during training. Specifically, we provide 2D masks for half of the parts and 2D bounding boxes for the remaining parts. This setting evaluates whether \modelname can seamlessly integrate heterogeneous strong priors and generalize to novel combinations of geometric conditions.

\textbf{Minimal input with part count only.}
We also consider the extreme case where users specify only the desired number of parts. Since FlexPart requires at least point-level spatial prompts, we automatically derive point prompts from the input image. Specifically, we extract high-resolution patch features $\mathbf{f}_{\mathrm{patch}}$ using DINOv2. To improve spatial consistency, we incorporate the 2D coordinates $\mathbf{x}_{\mathrm{patch}}$ of each patch to form a joint representation $\mathbf{v}_{\mathrm{patch}} = [\mathbf{f}_{\mathrm{patch}} \parallel \omega_{\mathrm{sp}} \mathbf{x}_{\mathrm{patch}}]$, where the spatial weight is set to $\omega_{\mathrm{sp}} = 5.0$. We then perform spectral clustering on this combined representation using $k=15$ nearest neighbors to obtain point prompts from the cluster centroids. In our preliminary experiments, we compared several clustering methods, including K-means and hierarchical clustering; however, spectral clustering consistently yielded the best performance in capturing semantically and spatially coherent part structures. This strategy enables effective part generation even from minimal input, and outperforms PartCrafter when only part counts are provided.

The quantitative results for these controlled input variations are summarized in Tab.~\ref{tab:input_analysis}. The results demonstrate a clear trade-off between the strength of geometric conditions and generation accuracy: stronger inputs provide more precise part localization, while weaker inputs offer greater flexibility with reduced supervision. Crucially, thanks to its unified multi-granularity design, \modelname consistently maintains reasonable performance across all input settings and generalizes well to noisy, partial, and mixed conditions. This highlights the advantage of a unified framework that seamlessly integrates heterogeneous user inputs for robust, practical interactive 3D part generation.

Complementary evaluation with human-provided point and box prompts is reported in \underline{\textbf{App.~\ref{app:human_prompts}}}.

\section{Evaluation with Human-Provided Prompts}
\label{app:human_prompts}

The controlled perturbations in \underline{\textbf{App.~\ref{User_Input_Variations}}} approximate common input errors. To complement them with actual human-provided prompts, we invite three annotators to mark their intended parts on the input images of 50 PartObjaverse-Tiny objects using points and 2D bounding boxes. Every evaluated input contains both prompt types; this study evaluates joint point-and-box input. Table~\ref{tab:human_prompts} reports the mean and standard deviation of the evaluation scores across the three annotators.

\begin{table}[t]
    \centering
    \small
    \caption{Human-provided prompts on 50 PartObjaverse-Tiny objects. All inputs contain both points and 2D boxes. Values are mean $\pm$ standard deviation across three annotators. IoU$_V$ and IoU$_B$ denote part-level voxel and bounding-box IoU, respectively.}
    \label{tab:human_prompts}
    \begin{tabular}{lccc}
        \toprule
        Prompt source & Part CD $\downarrow$ & IoU$_V$ $\uparrow$ & IoU$_B$ $\uparrow$ \\
        \midrule
        Human points + 2D boxes & $0.384\pm0.022$ & $0.131\pm0.010$ & $0.231\pm0.013$ \\
        \bottomrule
    \end{tabular}
\end{table}

We compute these metrics against the dataset parts after Hungarian matching. Since an annotator's intended decomposition may differ from the dataset annotation, the scores serve as reference measurements of agreement with the dataset, rather than strict measures of whether the user's intended decomposition is satisfied. The observed cross-annotator variation suggests limited sensitivity to the annotator within this study. This small-scale evaluation provides evidence for generation with human-provided spatial prompts, but does not measure interaction time or downstream usability. Complementary qualitative results with manually provided point or box prompts are shown in Fig.~\ref{fig:complex_topo}.

\section{Part Generation Efficiency}
\label{Appendix:efficiency}
\begin{figure}[t]
	\centering
		\includegraphics[width=0.9\linewidth]{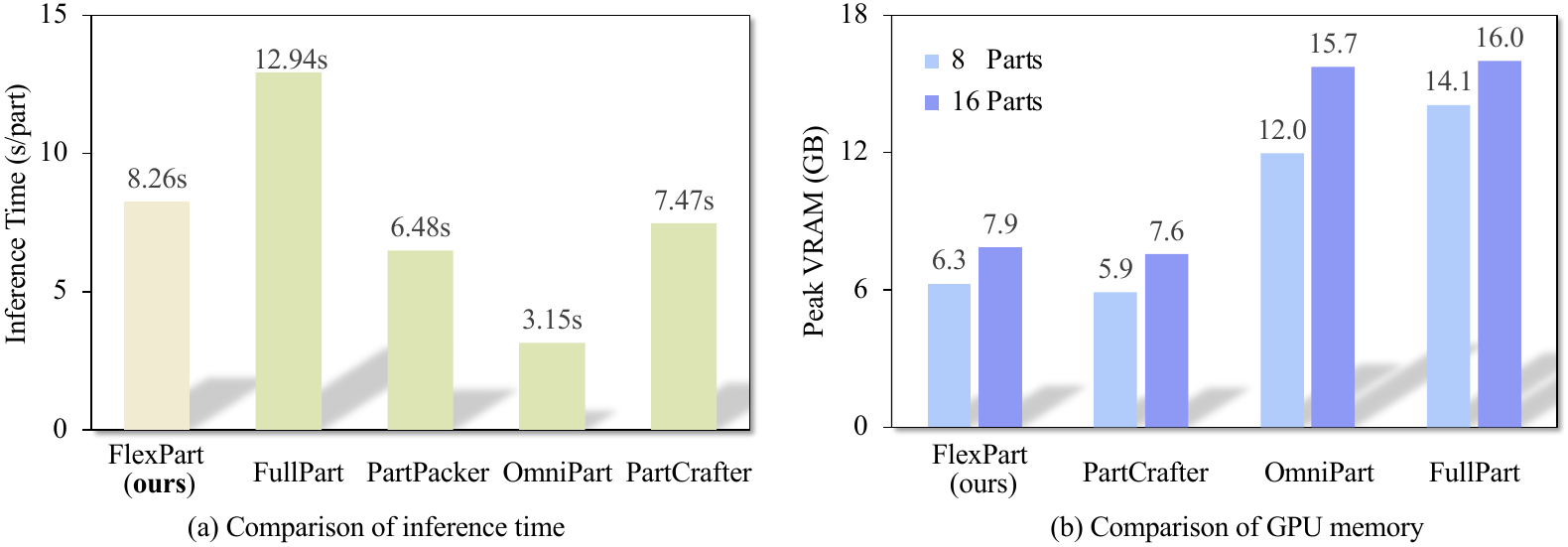}
		\caption{Efficiency comparison between \modelname and other methods.}
	\label{fig:efficiency}
        
\end{figure}

Beyond accuracy, we evaluate efficiency in terms of inference latency and memory usage.
Specifically, we measure the average time required to generate a single part, covering the entire process from condition encoding to final mesh extraction. Additionally, we analyze peak GPU memory (VRAM) consumption during object generation with varying numbers of parts. All evaluations are conducted on a single NVIDIA A800 GPU using random instances from the PartVerseXL validation set.

Regarding inference speed, Fig.~\ref{fig:efficiency}(a) illustrates that \modelname achieves competitive performance relative to existing baselines. Although our method shares a similar Diffusion Transformer backbone with PartCrafter~\cite{lin2025partcrafter}, we concatenate part identity embeddings with the latent code rather than adding them directly. This design choice introduces a slight computational overhead during attention operations, resulting in a marginal increase in inference latency. Compared to two-stage approaches like OmniPart~\cite{yang2025omnipart}, which accelerates generation via latent downsampling, our method operates at a higher resolution and thus requires more computation. \modelname strikes a balanced trade-off between generation quality and speed, leaving potential for further acceleration in future work.

In terms of memory consumption, as shown in Fig.~\ref{fig:efficiency}(b), \modelname maintains a modest peak VRAM footprint of $\sim$6GB to 8GB, which is comparable to the lightweight baseline PartCrafter~\cite{lin2025partcrafter}. In contrast, dense generation methods such as OmniPart~\cite{yang2025omnipart} and FullPart~\cite{ding2025fullpart} demand significantly higher resources, consuming nearly twice the memory under identical settings. Furthermore, our approach exhibits robust scalability with respect to object complexity. When doubling the target part count from 8 to 16, the peak memory usage increases only marginally. This footprint suggests that consumer GPUs with sufficient VRAM can support complex multi-part generation.

\graphicspath{{figures/}}

\section{Sensitivity of Point Prompts}
\label{Appendix:point_sensitivity}
\begin{figure}[t]
  \centering
  \includegraphics[width=\linewidth]{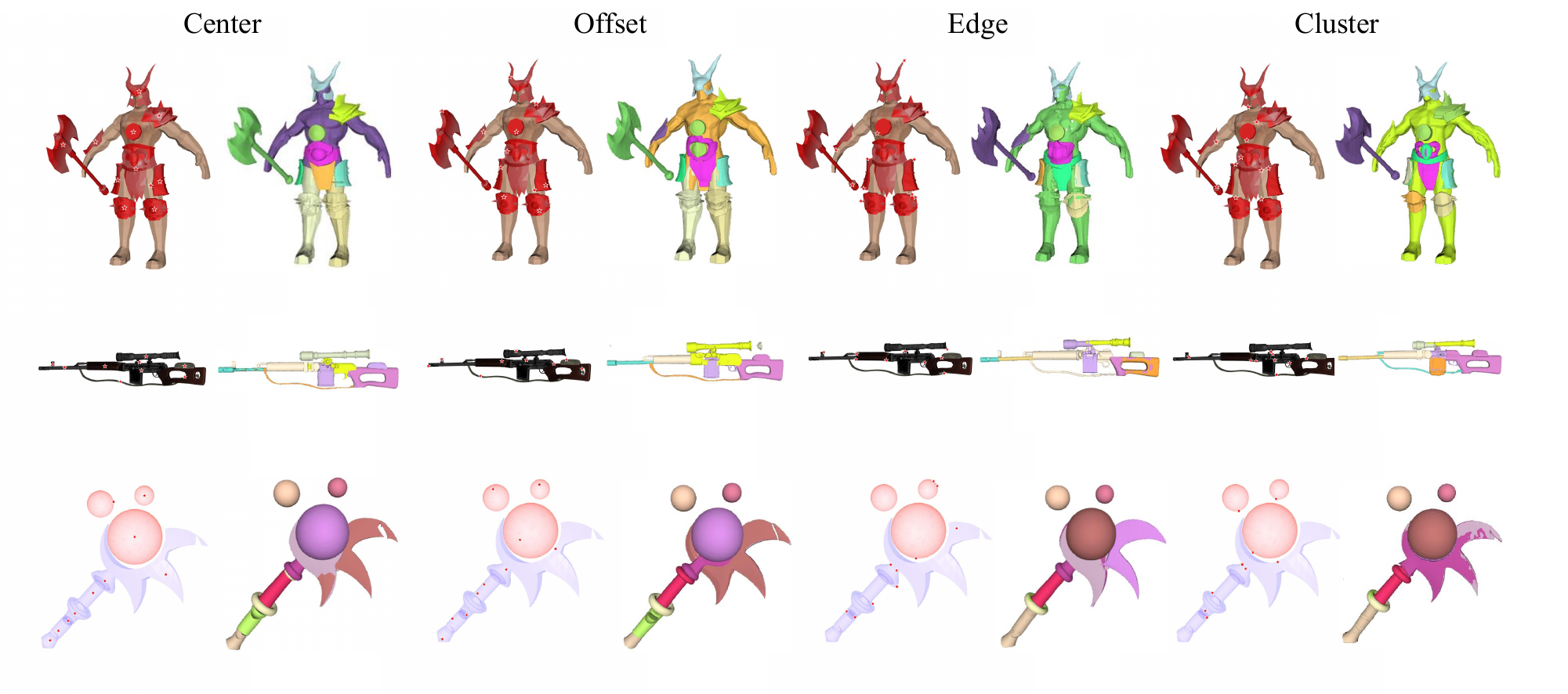}
  \caption {Qualitative analysis of point placement under four sampling strategies: Center, Offset, Edge, and Cluster.}
  
  \label{fig:sampling_qualitative}
  
\end{figure}

We investigate how point placement affects generation quality across input modalities, focusing on spatial ambiguity and the role of point prompts.

\textbf{Point Sampling Strategies.}
We consider four representative strategies to simulate diverse user behaviors and annotation noise: 
1). \textit{Center}, the centroid of the part mask; 
2). \textit{Edge}, a point sampled near the part boundary to induce maximal spatial ambiguity; 
3). \textit{Offset}, a randomly perturbed centroid to mimic typical user imprecision; 
4). \textit{Cluster}, the patch center nearest the object's geometric center.

\textbf{Quantitative Analysis.}
As shown in Tab.~\ref{tab:point_sensitivity}, the sensitivity to point placement strongly depends on the availability of additional geometric priors. 
When using \textit{Point Only} inputs, the model exhibits clear sensitivity to suboptimal point locations. In particular, sampling points near object boundaries (\textit{Edge}) leads to a substantial degradation across all metrics, with CD increasing from 0.3302 to 0.4190 and IoU$_V$ dropping from 0.1905 to 0.1342, indicating greater spatial ambiguity without supporting cues. 

In contrast, when stronger geometric conditions such as bounding boxes or masks are provided, the influence of point placement becomes significantly less pronounced. For example, under the \textit{Point + Box} setting, the variation across different strategies remains within a narrow range (e.g., CD varies only from 0.3036 to 0.3178), indicating stable performance. A similar trend is observed for \textit{Point + Mask}, where IoU$_V$ fluctuates marginally between 0.1962 and 0.2088, suggesting that geometric priors effectively dominate the spatial reasoning process.

\textbf{Qualitative Analysis.}
Consistent with the quantitative findings, our qualitative results shown in Fig.~\ref{fig:sampling_qualitative} visualize the generation outcomes under the four point sampling strategies in the \textit{Point Only} setting. The visual comparison clearly demonstrates how ambiguous point inputs, such as those near the edges, lead to a loss of geometric fidelity.

Overall, while \modelname is sensitive to point locations under point-only conditions, its sensitivity to point placement is substantially reduced when stronger geometric prompts are also provided.

\begin{table}[t]
\centering
\caption{Robustness analysis of point prompt placement across different input modalities. Metrics are evaluated on the PartVerseXL test set.}
\label{tab:point_sensitivity}
\scriptsize
\setlength{\tabcolsep}{8pt}
\begin{tabular}{llccc}
\toprule
\textbf{Input Type} & \textbf{Point Strategy} & \textbf{CD} $\downarrow$ & \textbf{IoU$_V$} $\uparrow$ & \textbf{IoU$_B$} $\uparrow$ \\
\midrule
\multirow{4}{*}{Point Only} 
 & Center  & \textbf{0.3302} & \textbf{0.1905} & \textbf{0.2513} \\
 & Edge    & 0.4190 & 0.1342 & 0.1697 \\
 & Offset  & 0.3476 & 0.1898 & 0.2455 \\
 & Cluster & 0.3803 & 0.1632 & 0.1905 \\
\midrule
\multirow{4}{*}{Point + Box} 
 & Center  & 0.3118 & \textbf{0.2074} & 0.2728 \\
 & Edge    & 0.3178 & 0.1996 & 0.2703 \\
 & Offset  & 0.3064 & 0.2029 & \textbf{0.2805} \\
 & Cluster & \textbf{0.3036} & 0.2062 & 0.2793 \\
\midrule
\multirow{4}{*}{Point + Mask} 
 & Center  & \textbf{0.2995} & 0.2030 & \textbf{0.2857} \\
 & Edge    & 0.3087 & 0.1962 & 0.2766 \\
 & Offset  & 0.3011 & \textbf{0.2088} & 0.2809 \\
 & Cluster & 0.3058 & 0.2042 & 0.2796 \\
\bottomrule
\end{tabular}
\end{table}

\graphicspath{{figures/}}

\begin{table}[t]
\centering
\caption{Comparison of direct point conditioning and mask-conditioned generation using SAM-derived or ground-truth masks. Direct point conditioning outperforms both SAM cascades on the reported part-level metrics.}
\label{tab:sam_comparison}
{
\scriptsize
\setlength{\tabcolsep}{8pt}
\begin{tabular}{llccc}
\toprule
\textbf{Setting} & \textbf{Condition} & \textbf{CD} $\downarrow$ & \textbf{IoU$_V$} $\uparrow$ & \textbf{IoU$_B$} $\uparrow$ \\
\midrule
SAM (Random) & Mask & 0.3498 & 0.1796 & 0.2375 \\
SAM (Center) & Mask & 0.3387 & 0.1731 & 0.2244 \\
\midrule
Direct point (ours) & Point & 0.3334 & 0.1902 & 0.2506 \\
\midrule
Ground Truth (Random) & Mask & \textbf{0.2938} & \textbf{0.2049} & \textbf{0.2840} \\
\bottomrule
\end{tabular}
}
\end{table}

\section{Analysis of SAM-based Cascaded Pipelines}
\label{Appendix:SAM_Analysis}
\begin{figure}[t]
  \centering
  \includegraphics[width=0.9\linewidth]{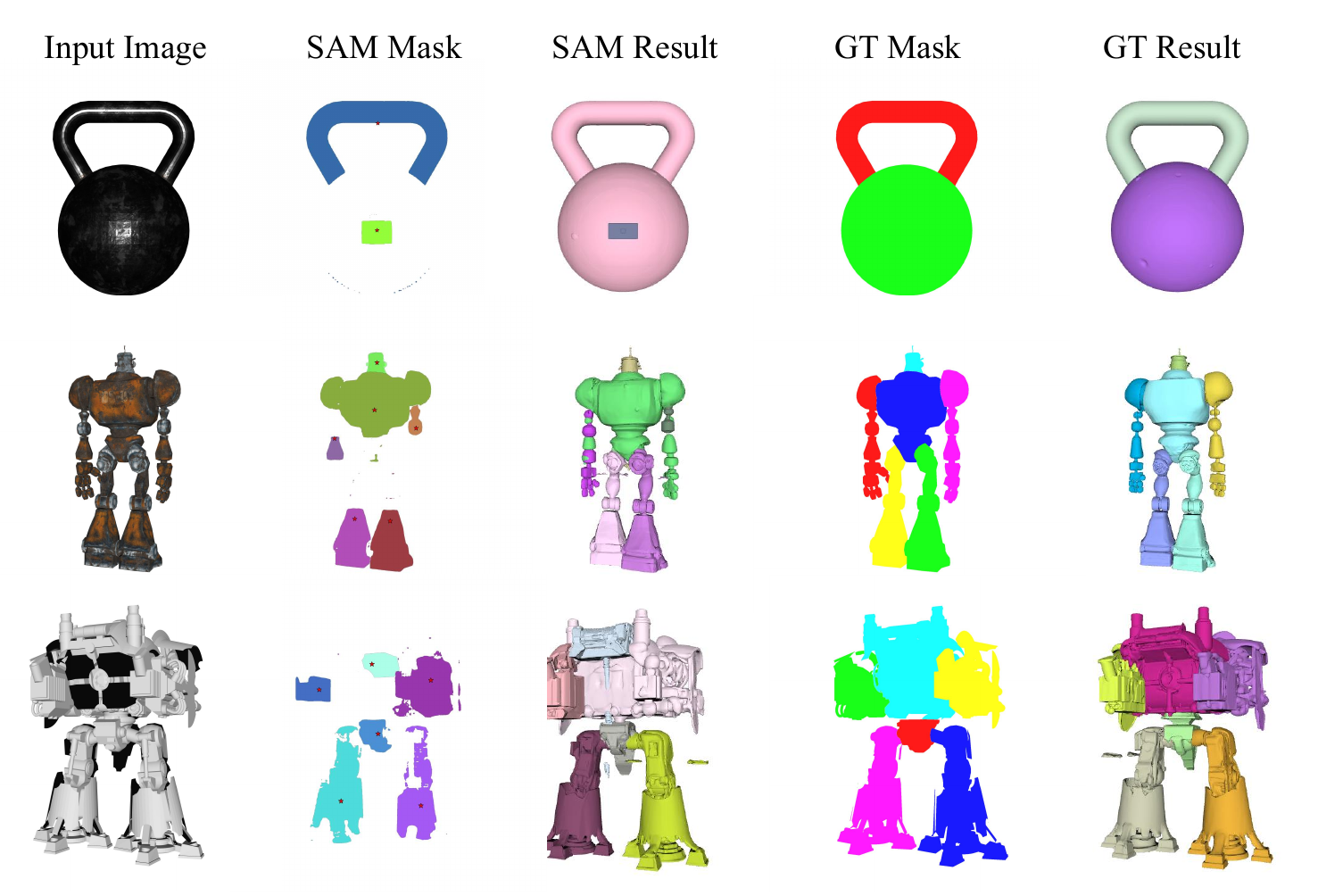}
  \caption{Qualitative evaluation of the SAM-based interface. We compare generation results under two input conditions across diverse examples. \textbf{SAM}: masks generated by SAM from a single point prompt sampled near the visual center. \textbf{GT}: ground-truth segmentation masks.}
  \label{fig:sam_qualitative}
  
\end{figure}
A seemingly plausible approach is to first convert sparse inputs (e.g., points) into dense masks using a 2D segmentation model such as Segment Anything (SAM)~\cite{kirillov2023segment}, followed by mask-conditioned 3D generation. We analyze this pipeline to understand whether such a decomposition can replace direct modeling of sparse conditions.

As shown in Tab.~\ref{tab:sam_comparison}, generation conditioned on SAM-derived masks performs worse than generation conditioned on ground-truth masks under both evaluated point sampling strategies. Even with center-based point sampling, CD degrades from 0.2938 to 0.3387, and IoU$_V$ drops from 0.2049 to 0.1731. Direct point conditioning achieves Part CD 0.3334, IoU$_V$ 0.1902, and IoU$_B$ 0.2506, outperforming both SAM-derived mask settings under the same evaluation protocol.

This performance gap may reflect a mismatch between image segmentation and the intended 3D part decomposition. As visualized in Fig.~\ref{fig:sam_qualitative}, given a sparse prompt, SAM typically segments a localized fragment rather than the complete functional part intended by the user. Incomplete intermediate masks can propagate segmentation errors into 3D generation.

In contrast, our end-to-end framework avoids this by directly modeling heterogeneous conditions. This joint modeling enables cross-modal synergy, allowing the network to learn robust semantics directly from sparse prompts. Furthermore, it naturally scales to structured 3D conditions like 3D bounding boxes, which cannot be elegantly reduced to 2D masks.

\graphicspath{{figures/}} 
\begin{figure}[t]
  \centering
  \includegraphics[width=0.85\linewidth]{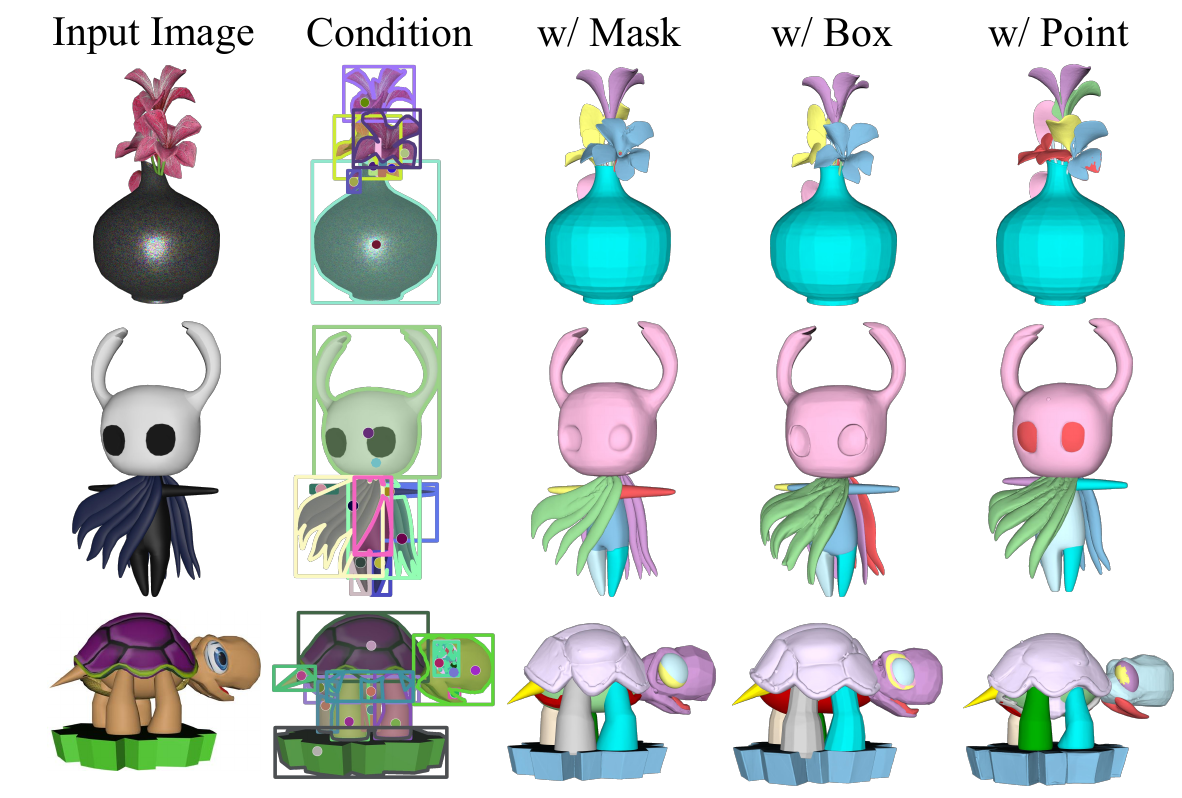}
  \caption{Additional qualitative results of \modelname across objects and geometric prompts.}
  \label{fig:more_results}
  
\end{figure}
\section{Additional Qualitative Results}
\begin{figure}[t]
  \centering
  \includegraphics[width=0.98\linewidth]{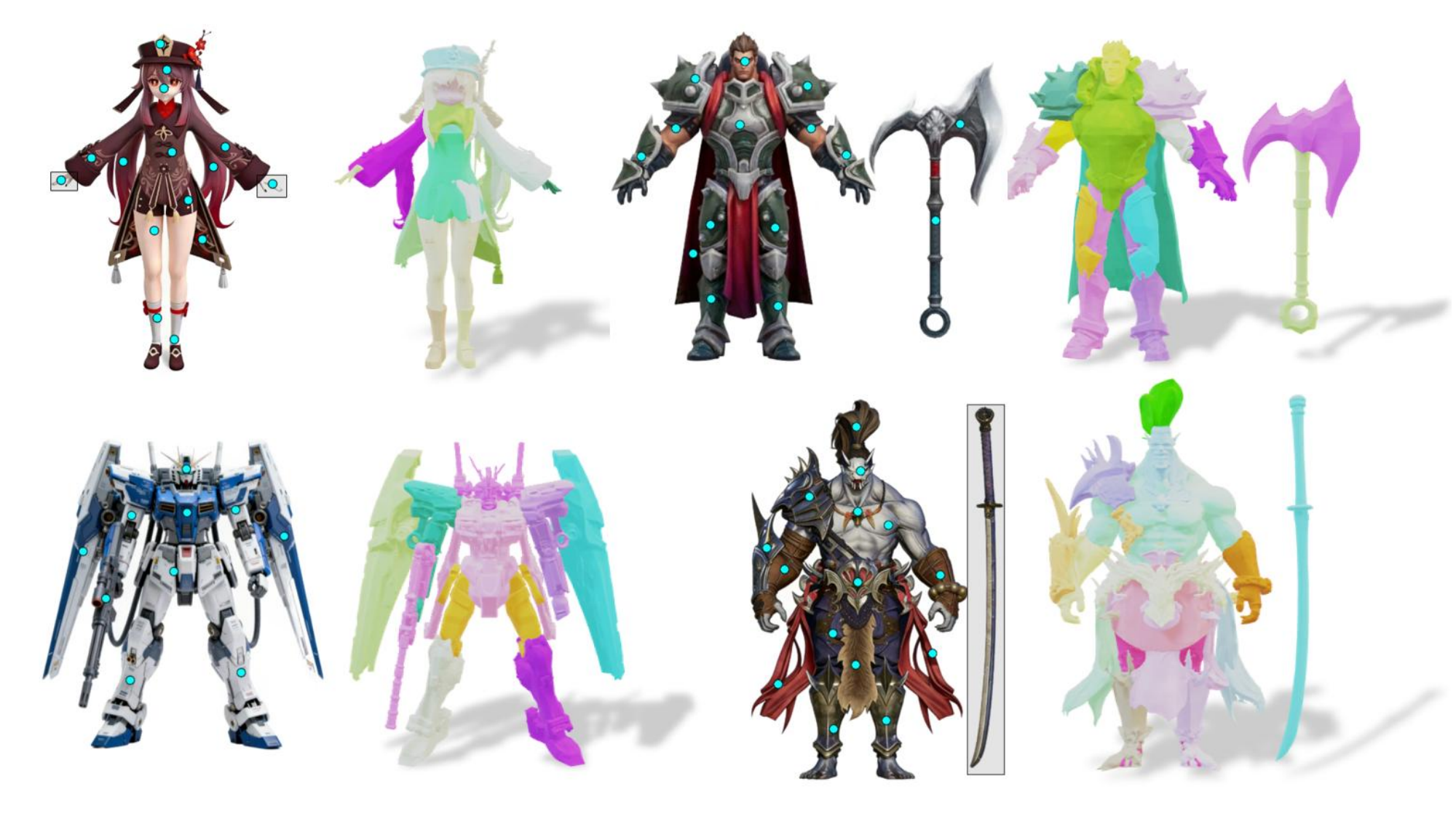}
  \caption{Generalization to complex topologies. Our method directly processes user-provided point or box prompts to generate 3D part assets.}
  \label{fig:complex_topo}
  
\end{figure}

To further substantiate the versatility of \modelname, we provide extended qualitative results in Fig.~\ref{fig:more_results} across a spectrum of geometric conditions. In addition, we present a new set of visualizations in Fig.~\ref{fig:complex_topo} focusing on objects with highly complex topological structures. By directly leveraging image conditioning alongside user-provided point or box prompts, our method successfully synthesizes high-quality, part-level 3D objects that preserve intricate geometric details.

These combined results highlight the core advantage of our unified framework, which empowers users to flexibly choose the most suitable prompt type for their specific needs. Regardless of the input modality, from lightweight points to dense masks, \modelname consistently produces high-fidelity meshes with clean topology and precise alignment to the provided spatial layouts, demonstrating robust generalization across diverse interaction modes.

\graphicspath{{figures/}} 



\end{document}